\pdfoutput=1
\RequirePackage{fix-cm}
\documentclass[twocolumn]{svjour3}          
\smartqed  

\usepackage{graphicx}

\usepackage{url}
\usepackage{xcolor}
\usepackage{hyperref}
\hypersetup{
     colorlinks=true,
     linkcolor=orange,
     filecolor=orange,
     citecolor=orange,      
     urlcolor=orange,
     }
\usepackage{color, colortbl}
\usepackage{amsfonts}
\usepackage{amsmath}
\usepackage{appendix}
\usepackage{algorithm, setspace}
\usepackage{algpseudocode}
\usepackage{csquotes}
\usepackage{amsfonts}
\usepackage{booktabs}
\usepackage{siunitx}
\usepackage{breqn}
\usepackage[normalem]{ulem}
\usepackage{tikz}
\usetikzlibrary{tikzmark}

\newcommand{\p}[1]{\medskip \noindent \textbf{{#1}.}}
\newcommand{\eq}[1]{Equation~(\ref{eq:#1})}
\newcommand{\fig}[1]{Figure~\ref{fig:#1}}

\journalname{Autonomous Robots}

\DeclareUnicodeCharacter{2212}{-}
\begin{document}

\title{CIVIL: Causal and Intuitive Visual Imitation Learning}


    \author{Yinlong Dai* \and Robert Ramirez Sanchez* \and Ryan Jeronimus \and Shahabedin Sagheb \and Cara M. Nunez \and Heramb Nemlekar \and \\ Dylan P. Losey}


\institute{Y. Dai, R. Sanchez, R. Jeronimus, S. Sagheb, and D. Losey are members of the Collaborative Robotics Lab, Mechanical Engineering Department, Virginia Tech.
\\
H. Nemlekar is a member of the Mechanical Engineering Department, California State University, Northridge. 
\\
C. Nunez is a member of the Mechanical Engineering Department, Cornell University. \\
Corresponding author's email:              \email{daiyinlong@vt.edu} 
}

\maketitle

\begin{abstract}
Today's robots attempt to learn new tasks by imitating human examples. 
These robots watch the human complete the task, and then try to match the actions taken by the human expert. 
However, this standard approach to visual imitation learning is fundamentally limited: the robot observes \textit{what} the human does, but not \textit{why} the human chooses those behaviors.
Without understanding which features of the system or environment factor into the human's decisions, robot learners often misinterpret the human's examples (e.g., the robot incorrectly thinks the human picked up a coffee cup because of the color of clutter in the background).
In practice, this results in causal confusion, inefficient learning, and robot policies that fail when the environment changes.
We therefore propose a shift in perspective: instead of asking human teachers just to show what actions the robot should take, we also enable humans to intuitively indicate \textit{why} they made those decisions (i.e., what features are critical for the desired task).
Under our paradigm human teachers attach markers to task-relevant objects and use natural language prompts to describe their state representation.
Our proposed algorithm, CIVIL, leverages this augmented demonstration data to filter the robot's visual observations and extract a feature representation that aligns with the human teacher.
CIVIL then applies these causal features to train a transformer-based policy that --- when tested on the robot --- is able to emulate human behaviors without being confused by visual distractors or irrelevant items.
Our simulations and real-world experiments demonstrate that robots trained with CIVIL learn both what actions to take and why to take those actions, resulting in better performance than state-of-the-art baselines.
From the human's perspective, our user study reveals that this new training paradigm actually reduces the total time required for the robot to learn the task, and also improves the robot's performance in previously unseen scenarios.
See videos at our project website: \url{https://civil2025.github.io}
\end{abstract}

\keywords{Visual Imitation Learning, State Representation, Few-Shot Learning}

\section{Introduction} \label{sec:intro}

\begin{figure*}[t]
	\begin{center}
        \includegraphics[width=\linewidth]{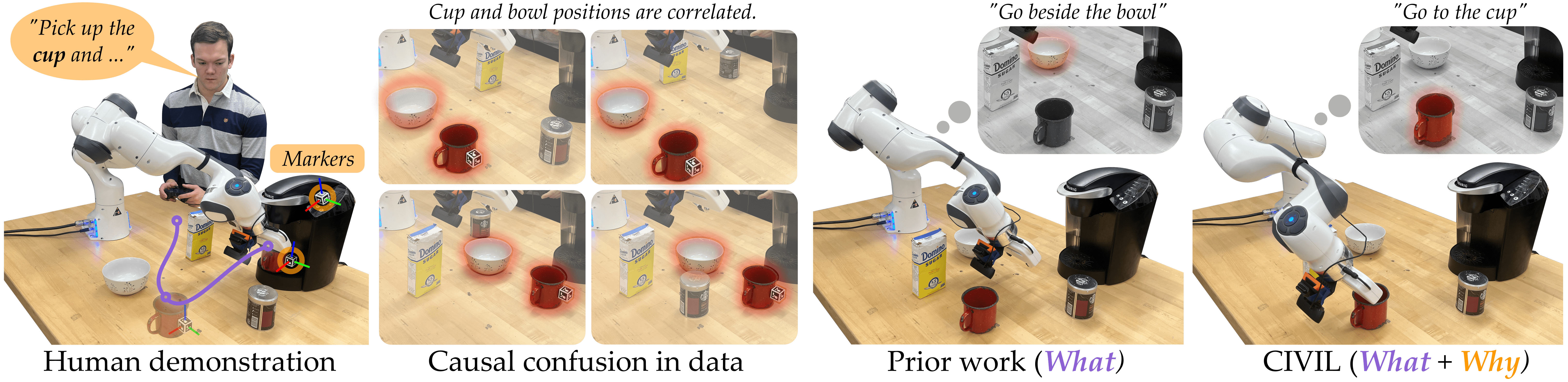}
		\caption{Human teaching a robot arm to prepare a cup of coffee. The robot must learn to grasp the cup and place it under a coffee machine based on visual observations. Within traditional approaches the human demonstrates \textit{what} actions to take, and the robot learns to emulate these demonstrated actions. However, this approach is inefficient because the robot is not taught \textit{why} the human chooses a specific behavior (i.e., what features of the environment factored into the human's decisions). Without this causal information that links features to actions the robot can misinterpret the human: for instance, if a bowl is always placed to the left of the cup during the demonstrations, the robot might learn to \textit{go beside the bowl} instead of \textit{go to the cup}. We hypothesize that robots can learn more efficient and robust control policies when the human teacher communicates the features behind their decisions (i.e., \textit{why} they are choosing the actions they demonstrate). CIVIL shifts imitation learning towards holistic demonstrations with physical markers and natural language instructions.}
		\label{fig:problem}
	\end{center}
    \vspace{-1em}
\end{figure*}

Imitation learning enables robots to learn new tasks by emulating the actions of a human expert. 
Consider a human teaching their robot arm to serve coffee (as shown in \fig{problem}). 
The human guides the robot through different stages of the task, including picking a cup off the kitchen counter and placing it under the coffee machine. 
To learn this task, the robot observes the scene with an onboard camera and records the actions demonstrated by the human teacher.
But these visual demonstrations only show the robot \textit{what} it should do, leaving the robot to figure out \textit{why} it should perform these actions (i.e., what aspects of the system and environment states factored into the human's decisions).

Understanding the state representation behind the human's actions is critical for adapting to new situations. 
For example, humans know that the coffee cup's position affects how it should be grasped; if the cup moves, humans will change their actions to match its new configuration. 
However, it is difficult for robots to infer this underlying feature solely from the demonstrated actions because their visual observations often contain excess information --- along with the cup, the robot's camera also sees other utensils and appliances on the kitchen counter.
These irrelevant details can create \textit{causal confusion} when they are correlated with the human's actions~\cite{de2019causal}. 
For instance, if the cup is always next to a bowl during the demonstrations, the robot may not understand the human's motive; should it reach for the cup or go to a position beside the bowl?
Put another way --- \textit{what features of the observed state are connected to the task, and which features are irrelevant clutter or spurious correlations?}

Existing research has focused on enabling robots to resolve this confusion on their own by making assumptions about task-relevant information. 
Current imitation learning methods try to extract the relevant details from the robot's observations by augmenting the data with random transformations~\cite{pari2021surprising,karamcheti2023language}, identifying known objects in the scene~\cite{zhu2023viola}, or using vision-language models pretrained on large datasets~\cite{sharma2023lossless,yang2024robot}. 
While these approaches help robots adapt their actions to expected variations of the task, they require a significant amount of data to truly uncover the human's reasoning.
For example, when we experimentally applied these baselines to the task in \fig{problem}, we found that the robot may incorrectly learn to focus on the bowl (instead of the coffee cup) because of misleading correlations in the training data.
This leads to robots that cannot make coffee when the bowl is removed.

To address this fundamental limitation we here re-frame the process of learning from human demonstrations. 
Rather than expecting robots to infer the correct causality based solely on human actions, we now extend imitation learning so that human teachers can intuitively reveal \textit{what} actions to take and \textit{why} to take those actions (i.e., what environment features guided the human's demonstrations). Our hypothesis is:
\begin{center} \vspace{-0.2em}
    \textit{Robots can learn more effectively when the human provides a smaller number of demonstrations while communicating the key features behind their actions.} \vspace{-0.2em}
\end{center}
We apply this hypothesis to create interfaces that humans can leverage to convey state representations during their demonstrations.
Specifically, we use a combination of physical markers and language instructions to give context to human demonstrations.
Human teachers place markers in the environment to highlight relevant objects, positions, and interactions that inform their actions (i.e., the human in \fig{problem} might mark the coffee cup and coffee machine).
Similarly, the human can provide natural language utterances to explain what they are doing or what they are focusing on during their demonstration (i.e., ``pick up the cup'').
The robot learner collects the demonstrated state and actions --- as in traditional approaches --- along with the new marker positions and language prompts.

These augmented demonstrations provide the robot with a more holistic understanding of the task and supplement its learning in two ways.
First, the robot leverages the marker and language cues to filter its extraneous observations and extract a low-dimensional feature representation that encodes human reasoning. 
Second, the robot learns a policy that maps these causal features to the demonstrated actions while remaining robust to unintended correlations and irrelevant visual data.
We refer to our resulting algorithm as
\textbf{CIVIL}: \textbf{C}ausal and \textbf{I}ntuitive \textbf{V}isual \textbf{I}mitation \textbf{L}earning.
Using CIVIL, humans can provide the robot with labeled data about the system and environment features (\textit{why}) that guided their demonstrationed actions (\textit{what}).  
Our results reveal that users perceive this immersive teaching protocol to be more intuitive and natural (i.e., how humans would teach other humans).
We also emphasize that gathering the additional marker and language data does not increase the overall teaching burden: instead, we find that users require fewer demonstrations and less total time to train the robot, and the resulting robot policy is more robust to new scenarios.

This work is a step towards robots that are able to correctly understand and perform tasks based on a few human demonstrations.
Overall, we make the following contributions\footnote{A preliminary version of this work was published at the IEEE International
Conference on Intelligent Robots and Systems \cite{sanchez2024recon}. As compared to \cite{sanchez2024recon}, this manuscript i) provides theoretical analysis justifying our proposed demonstration paradigm, ii) develops an end-to-end vision and language network that extracts causal features from human guidance, and iii) compares CIVIL to state-of-the-art alternatives while evaluating how humans interact with CIVIL.}:

\p{Analyzing Challenges in Visual Imitation Learning} 
We show why it is fundamentally challenging for robots to learn from high-dimensional and redundant observations, such as images from the robot's camera.
Using linear regression analysis, we first prove that humans must provide exponentially more examples as the dimensionality of observations increases.
We then illustrate why robots struggle to infer the human's reasoning and generalize to new scenarios when their observations contain spurious correlations.

\p{Introducing CIVIL}
To address these fundamental challenges, we enable humans to demonstrate tasks while also explaining their actions with physical markers and language instructions.
We present our CIVIL algorithm that leverages these inputs to train robots that i) extract causal features from their observations and then ii) map those features to task actions. 
Importantly, we only require markers and language commands during training. Once trained, the robot can perform the task autonomously without any supervision. 

\p{Comparing to State-of-the-Art Alternatives} 
We compare robots that act on the human-supervised features of CIVIL against multiple state-of-the-art baselines that let robots derive causality through self-supervision~\cite{pari2021surprising}, object detection~\cite{zhu2023viola}, and pre-trained vision-language models~\cite{radford2021learning}. 
Our experiments include simulations in CALVIN~\cite{mees2022calvin}, a benchmark for learning manipulation tasks, as well as real-world experiments with Franka robot arms. 
Robots trained on CIVIL are more successful in performing the tasks than the baselines, especially when tested on unseen task instances.

\p{Evaluating with Real Users}
We conduct experiments where real users leverage our CIVIL protocol to train the robot arm.
We focus on the user's subjective perception of the demonstration process, as well as the robot's objective performance when trained on user data.
Our results suggest that users find it easy and intuitive to leverage markers and language during demonstrations, and --- when giving the same amount of time for providing demonstrations --- robots trained with CIVIL learn to perform the task more proficiently.

\section{Related Work} \label{sec:related}

Our work explores visual imitation learning for robot manipulation tasks.
Below we summarize this field, while focusing on existing methods that enable the human teacher to augment their demonstrations.

\subsection{Visual Imitation Learning} \label{sec:R1}

When imitating humans, the robot learns a policy that maps its observations to the actions demonstrated by a human expert.
We expect robots to learn this expert policy from a few demonstrations and then transfer it to other, potentially unseen variations of the task~\cite{osa2018algorithmic,mandi2022towards}. 
But when the observations are high-dimensional and contain extraneous information, it can be difficult for robots to infer which parts of these observations actually affect the task performance~\cite{de2019causal}. 
For example, the robot in \fig{problem} may not know which objects to focus on when making coffee on a cluttered kitchen counter. 

To resolve this confusion, robots can encode their observations into a low-dimensional \textit{feature representation} that only retains essential information --- such as the position of the cup --- and ignores irrelevant details like lighting changes and background objects~\cite{xie2024decomposing}. 
Existing approaches let robots derive these features on their own by making assumptions about the extraneous aspects~\cite{pari2021surprising,radosavovic2023real,pfrommer2023initial,sharma2023lossless}, or simply focusing on the known objects in the scene~\cite{qin2022dexmv,zhu2023viola,jonnavittula2025view}.

For instance, the robot can generate alternative views of the images taken from its camera by applying transformations like color distortions and random cropping~\cite{chen2020simple,grill2020bootstrap,he2022masked}, and then train an encoder to map these transformed views into the same features as the original, unmodified images.
This helps the robot learn a feature representation that is invariant to noisy transformations.
Alternatively, the robot can use existing vision models to detect known objects in its view and train its policy on features derived from the segmented images of these objects~\cite{zhu2023viola}.
This approach encourages the robot to disregard background details like the appearance of the kitchen or the lighting of the room.

While these unsupervised approaches make the robot robust to distractors like lighting and background, they rely on the robot to implicitly infer the relevant features (e.g., the cup's position) from human demonstrations. 
This slows the learning process --- humans need to provide demonstrations in diverse scenarios to facilitate causal inference~\cite{bica2021invariant} --- and can also be counterproductive when the assumed variations deviate from the human's reasoning~\cite{grill2020bootstrap}. 
For example, if a user wants the robot to interact with objects of a specific color or focus on some background cues, training with images that vary in color or exclude the background can further confuse the robot.

Hence, in this work we enable humans to explicitly convey their underlying feature representations to the robot.
We anticipate that communicating the reasoning behind human actions will mitigate causal confusion and accelerate learning. 
Accordingly, we next discuss prior works that have explored how humans can intuitively reveal their intentions to robot learners.

\subsection{Learning Human Representations} \label{sec:R2}

Robots can learn more efficiently and generalize better to unseen scenarios when their representations are aligned with human reasoning~\cite{bobu2024aligning}.
To achieve this alignment, humans need to share further insights into their decision-making while demonstrating the task.
Earlier works have proposed obtaining representations by asking humans to select the task-relevant factors from a pre-defined list~\cite{cakmak2012designing,basu2018learning,nemlekar2023transfer}, label the features for examples in the training data~\cite{sripathy2022teaching,schrum2024maveric}, and provide demonstrations that trace the gradient of a relevant feature~\cite{bobu2022inducing}.
However, these approaches are either cognitively demanding because users find it difficult to quantify feature values~\cite{koppol2021interaction}, or physically taxing due to the need for additional feature-specific demonstrations.
To feasibly obtain this information in practice, it is important to leverage natural and intuitive communication channels that can be seamlessly integrated into the robot's training process~\cite{habibian2024survey}.
Therefore, recent work has focused on pairing demonstrations with natural language prompts~\cite{sharma2023lossless,mees2022matters,jang2022bc,yu2023using,ma2024actra,brohan2023rt,team2024octo,yang2024robot}, and introducing intuitive sensors and interfaces to collect additional human inputs~\cite{song2020grasping,young2021visual,song2023data,wei2024wearable,quintero2018robot,luebbers2021arc,saran2021efficiently,biswas2024gaze,nemlekar2024pecan}.

Humans can organically explain their actions using natural language. For example, when demonstrating how to make coffee, users may say ``pick up the cup'' and then ``place it under the coffee machine.'' Prior works have shown that robots can utilize these prompts to improve their representations in multiple ways. Many previous approaches encode language descriptions into feature vectors and pair them with visual features to provide more context for the robot's policy~\cite{mees2022matters,jang2022bc,yu2023using,ma2024actra,brohan2023rt,team2024octo}. 
Some works use language to supervise how features are extracted from robot observations by using contrastive learning, as in CLIP~\cite{radford2021learning}, or by conditioning their visual encoder~\cite{karamcheti2023language}.
Lastly, instead of learning the features from scratch, we can take pretrained vision-language models and fine-tune them on demonstrations of the task~\cite{sharma2023lossless,yang2024robot}.
In our work, instead of using language to contextualize the robot's features or policy, we leverage language to filter the robot's observations --- highlighting relevant objects in the scene and removing irrelevant details that can confuse the learner.

Although humans can explain parts of their thinking using natural language, not every aspect of a task can be easily put into words, e.g., subconscious visual cues or complex motion constraints.
Such details can be communicated more intuitively through specialized instruments. 
For instance, humans can cheaply convey rich motion information using hand-held grasping tools~\cite{song2020grasping,young2021visual}, optical trackers~\cite{song2023data}, and wearable tactile gloves~\cite{wei2024wearable}. Humans can also utilize augmented reality interfaces to specify keyframes and motion constraints~\cite{quintero2018robot,luebbers2021arc}. Alternatively, the robot can track human gaze and focus on the same regions of its observations as the expert user~\cite{saran2021efficiently,biswas2024gaze}.
We explored this option in our preliminary work \cite{sanchez2024recon}, where humans used  Bluetooth sensors to locate relevant objects in the environment; similarly, \cite{zhu2023learning} introduced an interface for humans to mark these objects on images of the scene~\cite{zhu2023learning}.
However, we find that physical markers alone are not sufficient to capture critical features.
These markers might indicate where the robot should focus (e.g., ``look at the light bulb''), but not what aspects to focus on (e.g., ``check if the light is on or off'').

Our work finds a balance between instrumented and natural human inputs. We use a combination of physical markers and language descriptions to specify relevant poses and objects that humans consider when taking actions. Our approach leverages these inputs to encode the robot's visual observations into a feature representation that is aligned with human reasoning. Unlike previous approaches, \textit{we only require additional inputs during training}. Once the robot learns the correct representation, it can autonomously perform new variations of the task without needing markers or language prompts.

\section{Problem Statement} \label{sec:problem}

We consider settings where a robot arm is learning a task from human demonstrations. 
When teaching a new task, the human teleoperates or kinesthetically moves the arm through a few instances of that task. For example, the human may show how a coffee cup can be picked up from different locations on the kitchen counter.

\p{Robot}
As the human demonstrates the task, the robot records its states $x\in \mathbb{R}^{m}$ (e.g., joint angles), actions $u \in \mathbb{R}^{m}$ (e.g., joint velocities), and observations $y \in \mathbb{R}^{n}$ (e.g., images taken from onboard and static cameras). While $x$ only represents the arm's proprioceptive state, $y$ also captures information about the surrounding environment. Overall, the human provides a dataset $D$ of $(x, y, u)$ tuples. The robot's goal is to leverage this dataset to learn a control policy $\pi_{\theta}$ that maps the states $x$ and observations $y$ to the demonstrated actions $u$:
\begin{equation}
  \pi_{\theta}(x, y) = u \quad \forall (x, y, u) \in D  
\end{equation}
The policy parameters $\theta$ determine \textit{what} actions the robot chooses for a given state and observation.

\p{Features}
The robot's observations are high-dimensional and contain both relevant information for learning the task and extraneous details that should be ignored. For instance, along with the cup that we want the robot to grasp, it could also see a bowl and other kitchen appliances on the counter.
The robot does not know which parts of these observations are relevant \textit{a priori}. 
We represent the task-relevant information as a compact feature vector $\phi^{*} \in \mathbb{R}^{d}$, where the feature dimension $d$ is less than the dimensionality $n$ of the observations.
In our example, $\phi^{*}$ contains the cup's position and orientation but excludes the bowl and other irrelevant items on the kitchen counter. 
Our work focuses on extracting the relevant features from the human's demonstrations so that we can map high-dimensional states into compact feature vectors.

\p{Human} Unlike the robot arm, humans know the task-relevant aspects and can extract the associated features from the high-dimensional observations through a feature function $f$.
\begin{equation}
  f_{\psi^{*}}(x, y) = \phi^{*}  \label{eq:human_feature}
\end{equation}
The parameters $\psi^{*}$ determine how humans map the complex observations to the relevant features. Without loss of generality, we assume that humans only act based on these features (e.g., the human will not focus on the bowl's position when reaching for the cup), and so the human's policy is a function of the relevant features $\phi^{*}$.
\begin{equation}
  \pi_{\theta^{*}}(x, \phi^{*}) = u  \label{eq:human_policy}
\end{equation}
In the above $\theta^{*}$ are the true parameters of the policy that the human wants to teach the robot.
Intuitively, the policy parameters $\theta^{*}$ dictate \textit{what} actions the human will take and the features $\phi^{*}$ determine \textit{why} the human chooses that action for a given robot state and observation (i.e., $\phi^*$ provides a feature state representation for the desired task).

Ideally, the control policy learned by the robot arm should produce the same actions as the human expert. 
In what follows, we discuss two key challenges in learning such a policy from visual observations given a limited amount of training data $D$. First, we highlight the importance of encoding the robot's observations into low-dimensional features (similar to those of the human expert) in order to improve learning efficiency. Second, we illustrate why it is difficult for robots to learn policies that can generalize to new task instances when their observations contain correlated visual elements.

\subsection{Using Low-Dimensional Features to Accelerate Learning} \label{sec:accelerate}

We first analyze the challenge of efficiently learning from high-dimensional observations like RGB images. More specifically, we show that the data required to learn the task increases exponentially as we increase the dimensionality of the inputs to the robot's policy. 
To formalize this problem, we consider a linear regression example where the robot has a dataset that contains $N$ samples of states $x \in \mathbb{R}^{m}$, observations $y \in \mathbb{R}^{n}$, and demonstrated actions $u \in \mathbb{R}^{m}$.
We assume that the robot encodes the states and observations into features $\phi \in \mathbb{R}^{d}$ using an encoder matrix $\Psi \in \mathbb{R}^{(m+n)\times d}$:
\begin{align*}
    \Phi = [X Y]\Psi
\end{align*}
where $X \in \mathbb{R}^{N\times m}$ and $Y \in \mathbb{R}^{N\times n}$ are the matrices formed by stacking the states $x$ and observations $y$ in the dataset, and $\Phi \in \mathbb{R}^{N\times d}$ is a matrix of corresponding features $\phi$.
For now, we provide the robot with a pretrained encoder $\Psi$ which extracts features that are sufficient for learning the task.

The robot's goal is to learn a matrix of policy parameters $\theta \in \mathbb{R}^{d \times m}$ that maps the features to actions $U \in \mathbb{R}^{N \times m}$:
\begin{align*}
    U = \Phi \theta +\epsilon
\end{align*}
The term $\epsilon$ accounts for any errors made by the human teacher when demonstrating the task actions.
Given this problem formulation, we now establish how the dimensionality $d$ of the features affects the number of samples $N$ required to learn the policy parameters $\theta$. Specifically, under the standard assumptions listed below, we prove that:

\p{Proposition 1}
When the demonstrated actions $u$ have a zero-mean Gaussian noise $\epsilon$, and the features $\phi$ input to the robot's policy are normally distributed, the amount of data $N$ required to learn the parameters $\hat{\theta}$ of a linear policy varies exponentially with the dimensionality $d$ of the features.

\begin{proof}
We assume that the actions demonstrated by the human have a zero-mean Gaussian noise:
\begin{align*}
    \epsilon \sim \mathcal{N} (0, \sigma_{\epsilon}^{2} I_{m})
\end{align*}
When the error follows a normal distribution, the ordinary least squares estimator $\hat{\theta}$ is also normally distributed~\cite{amemiya1985advanced}:
\begin{align*}
    \hat{\theta} \sim \mathcal{N} (\theta, \sigma_{\epsilon}^{2} (\Phi^{T} \Phi)^{-1})
\end{align*}
The variance in the estimated parameters corresponds to the uncertainty in converging to the human's policy. We quantify this uncertainty using the continuous Shannon entropy of the estimator's distribution~\cite{ahmed1989entropy}:
\begin{align*}
    h(f_{\hat{\theta}}) = \frac{d}{2} + \frac{d}{2}\ln2\pi + \frac{1}{2}\ln|\Sigma_{\hat{\theta}}|
\end{align*}
Here $d$ is the dimensionality of the features. We will now simplify the covariance term $\Sigma_{\hat{\theta}}$ to verify how this uncertainty scales with the feature dimensions. We start by writing $\Phi^{T} \Phi$ as a sum of the outer product of the feature vectors $\phi_{i} \in \Phi$:
\begin{align*}
    \Phi^{T} \Phi = \sum_{i=1}^{N} \phi_{i} \otimes \phi_{i}
\end{align*}
From the definition of variance~\cite{casella2024statistical}, the expected value of the outer product can be written in terms of the mean and variance of the feature vectors:
\begin{align*}
    \mathbb{E}\left[\sum_{i=1}^{N} \phi_{i}\phi_{i}^{T}\right] = \sum_{i=1}^{N}\left( \Sigma_{\phi} + \mu_{\phi}\mu_{\phi}^{T}\right)
\end{align*}
To simplify this further, we assume that the feature vectors are normally distributed such that $\phi_{i} \sim (0, \sigma_{\phi}^{2} I_{d})$. 
Note that this is a common assumption in previous approaches that use Variational Autoencoders (VAEs)~\cite{pinheiro2021variational} to encode image data. Equipped with this assumption, we can now write the expected value of the outer product as $\mathbb{E}[\Phi^{T}\Phi] = N\sigma_{\phi}^{2} I_{d}$ and substitute it back into the covariance of the estimator distribution: 
\begin{align*}
    \Sigma_{\hat{\theta}} = \sigma_{\epsilon}^{2} (N\sigma_{\phi}^{2})^{-1}I_{d}
\end{align*}
Finally, we take the determinant of the covariance matrix and express the entropy over the estimated parameters as:
\begin{equation}
    h(f_{\hat{\theta}}) \propto d \cdot \ln \left(\frac{1}{N} \cdot \frac{\sigma_{\epsilon}^{2}}{\sigma_{\phi}^{2}}\right)
\end{equation}
From this result, we observe that uncertainty in the learned parameters decreases logarithmically with the number of data samples $N$ but increases linearly with the number of feature dimensions $d$.
In other words, as we decrease the dimensionality of input features, humans would need to provide exponentially fewer data samples to converge to the human's policy. \qed
\end{proof}

Proposition 1 illustrates the importance of mapping the robot's high-dimensional observations into a minimal feature representation. 
But what is the right representation?
Thus far we have assumed that the robot has access to a feature function $\psi$ that extracts sufficient information for learning the task.
In the next subsection, we will show that there can be many feature representations that are sufficient for imitating the actions in the training data but do not align with the human's reasoning $\phi^{*}$.
As a result, these alternate feature representations are susceptible to covariate shift and fail when the robot encounters new states at test time.

\subsection{Causal Confusion in Visual Imitation Learning} \label{sec:causal_confusion}

When the robot does not have any prior knowledge of the task-relevant features, we can simultaneously train a feature function $f_{\psi}(x, y) = \phi$ and policy $\pi_{\theta}(x, \phi) = u$ on samples from the training dataset $D$:
\begin{equation}
  \pi_{\theta}(x, f_{\psi}(x, y)) = u \quad \forall (x, y, u) \in D  
\end{equation}
However, this does not guarantee that the features learned by the robot will match the task-relevant features $\phi^{*}$.
For instance, when teaching the robot to make coffee, imagine that the cup is always placed next to a bowl during training. While the human knows that only the cup is important, the robot may mistakenly learn to extract the bowl's pose (irrelevant features) or infer the cup's pose by observing the bowl instead (spurious correlations). 
Despite this incorrect mapping, the robot could learn a policy that successfully grasps the cup in all training instances because of its positional relationship with the bowl. 

More generally, these correlations create \textit{causal confusion}~\cite{de2019causal} when applying the learned feature function in unseen scenarios. 
To demonstrate this formally, we return to the linear regression problem. We now assume that the encoder matrix $\psi$ is unknown and combine it with the policy parameters to create a single weight matrix $W = \psi\theta$:
\begin{align*}
    U = [X Y] \psi\theta + \epsilon = [X Y] W + \epsilon
\end{align*}
In an ideal scenario, there is no noise ($\epsilon \rightarrow 0$) in the human's demonstrations, and the input matrix $[X Y]$ has full rank. This allows us to obtain an exact least-squares solution:
\begin{align*}
    \hat{W} = [X Y]^{\dag}U
\end{align*}
Here $[X Y]^{\dag}$ denotes the pseudo-inverse of $[X Y]$. While there are infinite ways to factorize $\hat{W}$ into the components $\hat{\psi}$ and $\hat{\theta}$, because we have a unique $\hat{W}$, all of these choices are equivalent to the human's true weights $W^{*}$:
\begin{align*}
    \hat{W} = \hat{\psi}\hat{\theta} = \psi^{*}\theta^{*} = W^{*}
\end{align*}

By contrast, when there is non-zero correlation between the input dimensions --- e.g., manipulating the cup that is next to a bowl --- the input matrix $[X Y]$ will have a non-trivial null space, resulting in an infinite number of solutions for $\hat{W}$:
\begin{align*}
    \hat{W} = W^{*} + V
\end{align*}
Here $V$ is any matrix in the null space of $[X Y]$ that satisfies $[X Y]V = 0$. 
This means that the weights learned by the robot \textit{will not match the true weights}, $\hat{W} \neq W^{*}$, except in the special case when $V=0$.

This difference in learned weights will not affect the robot's performance if the correlations in the training data are also present at test time. In this case, the test inputs $[X Y]_{test}$ will have the same null space as the training data:
\begin{align*}
    [X Y]_{test}V = 0
\end{align*}
As a result, $\hat{W}$ will produce the same actions as $W$.
However, if the correlations in the inputs \textit{change} at test time, then the learned weights $\hat{W}$ will not produce the same actions as $W^{*}$ for any non-trivial $V$. For instance, if the positions of the bowl and cup are no longer related during testing, the test inputs $[X, Y]_{test}$ will have full rank. This means that $V$ will not belong to the null space of $[X, Y]_{test}$:
\begin{align*}
    [X Y]_{test}V \neq 0
\end{align*}
As such, the actions predicted by the robot will differ from the true actions given by $W^{*}$:
\begin{equation}
    U_{test} = [X Y]_{test}W^{*} \neq [X Y]_{test}\hat{W}
\end{equation}
Overall, this result demonstrates that when the robot's observations contain unwanted correlations, the robot can still learn \textit{what} actions to take during training but it will not understand \textit{why} to take those actions. Because of this fundamental misalignment the learned policy may not generalize to new scenarios that differ from the training distribution.

\subsection{Problem Summary} \label{sec:P3}

In this section we showed that the amount of demonstration data required to train the robot policy increases exponentially with the dimensionality of the input states and observations. 
This slows down learning for robots that take actions based on dense inputs like camera images.
We can improve the learning efficiency by encoding robot observations into a compact feature representation. Unfortunately, if the observations contain misleading correlations, the encoded features will fail to correctly explain the human's actions --- regardless of how many demonstrations the human provides.

When correlations are present in the training dataset the robot has no way of determining causality. 
Instead of pushing this fundamental limitation entirely to the robot, we will enable humans to explicitly convey relevant visual cues and features during training. This additional information can help robots filter the spurious correlations in their observations and extract compact features that causally influence human actions.
In the following section we present our approach for obtaining these additional inputs from humans and training robots to 
mimic the human decision-making process (i.e., imitating \textit{what} the human does and \textit{why} they choose those actions).
\section{Causal and Intuitive \\ Visual Imitation Learning} \label{sec:method}

\begin{figure}[t]
	\begin{center}
 		\includegraphics[width=\linewidth]{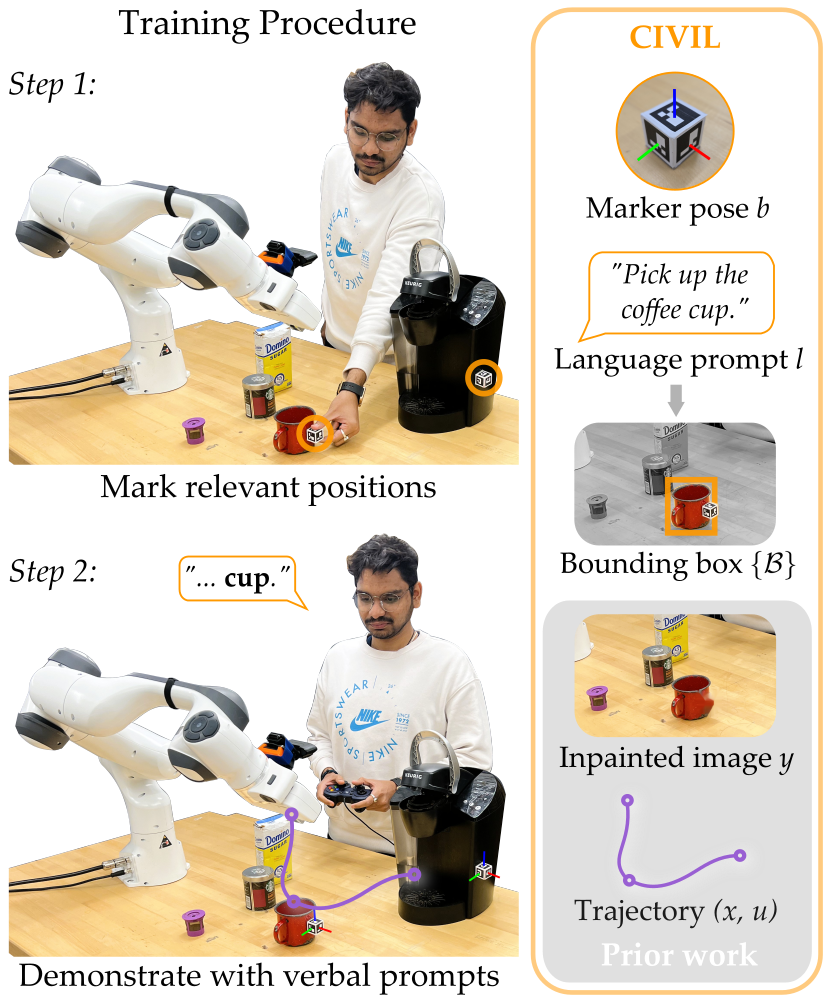}
		\caption{Augmented data collection procedure for CIVIL. In Step 1, we enable humans to mark task-relevant positions (e.g., the coffee maker) with ArUco markers. In Step 2, as the human demonstrates the task they can provide natural language prompts that mention task-relevant objects (e.g., the cup). The resulting dataset for offline learning includes states $x$, images $y$, actions $u$, marker data $b$, and language prompts $l$.
        After providing data, the human removes the markers from the environment, and the robot processes its images to inpaint those markers so that they are not required at test time.
        }
		\label{fig:method}
	\end{center}
    \vspace{-1em}
\end{figure}

We want robots to efficiently learn new tasks from human demonstrations and generalize the learned behavior to unseen task instances.
In the previous section we showed that understanding compact features is critical to efficient learning, but merely imitating human actions is not always sufficient to recover these features.
To address this problem, we here re-frame how humans provide demonstrations to include both showing the desired behavior (\textit{what}) and also highlighting the features that influence their behavior (\textit{why}).
We recognize that humans understand what aspects of the task are important to their decision-making process, and human teachers can label the task-relevant features $\phi^{*} = f_{\psi^{*}}(x, y)$ from visual observations (see \fig{method}).
Our proposed CIVIL algorithm then synthesizes the augmented demonstrations to perform offline visual imitation learning and recover the desired task.

In Section~\ref{sec:instruments} we first equip humans with \textit{instruments} that enable them to intuitively communicate information about the relevant features (i.e., $\phi^{*}$) and which parts of the observations they consider when extracting these features (i.e, $\psi^{*}$).
Next, in Section~\ref{sec:model}, we describe our network architecture for extracting features from robot observations and mapping them to corresponding robot actions.
We apply this architecture in Section~\ref{sec:supervised} to develop the CIVIL algorithm which synthesizes data collected from our instruments to align the robot's features with the human's reasoning. 
Finally, in Section~\ref{sec:M4} we provide implementation details.
A key contribution of our approach is that the robot does not need instruments after training and can perform the task autonomously at test time based only on visual observations.

\begin{figure*}[t]
	\begin{center}
 		\includegraphics[width=\linewidth]{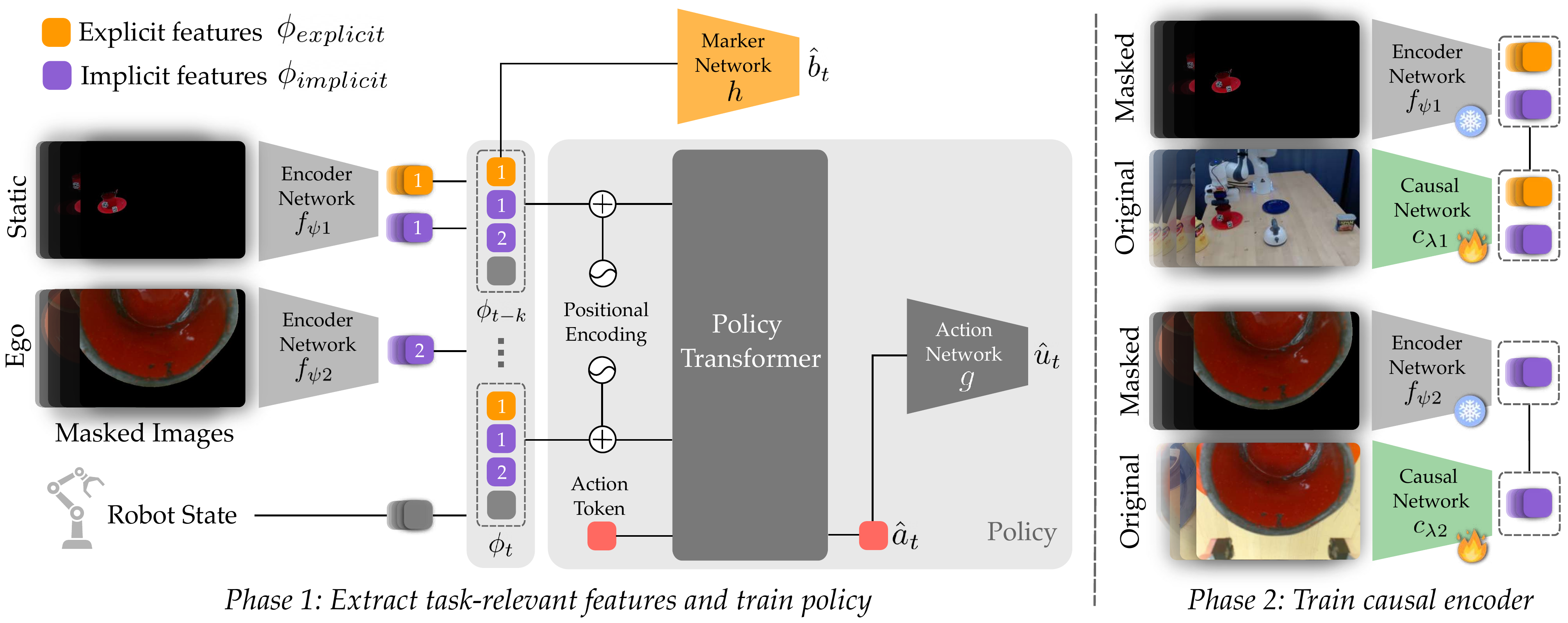}
		\caption{Network architecture of CIVIL. The model consists of encoder networks that map environment observations (images) to a compact feature representation $\phi$, and a policy transformer that takes a sequence of robot states and features as input and predicts the task action. The training of our model is split into two phases. (Left) In the first phase we supervise a subset of the features using a marker network $h$ to \textit{explicitly} encode the relevant poses $b$ marked by the human expert. At the same time, we train the remaining features to \textit{implicitly} capture other task-relevant information by masking the input images to highlight the relevant objects conveyed by the human through natural language instructions $l$.
        The features are trained together with the policy transformer by optimizing a dual loss function that aligns the robot's representation with human reasoning (the \textit{why}) and minimizes the error between predicted and ground truth actions (the \textit{what}). (Right) In the second phase we freeze the encoder network and policy network, and train a causal network $c$ to map the original images to the same features as those learned by the robot from the masked images in the first phase. This step ensures that the robot can extract the task-relevant features without needing the human to place markers or provide language prompts at runtime.}
		\label{fig:network}
	\end{center}
        \vspace{-1em}
\end{figure*}

\subsection{Obtaining Task-Relevant Information from Humans}\label{sec:instruments}

We envision two channels for humans to intuitively explain their thinking when demonstrating tasks: i) conveying the features they extract, and ii) highlighting the visual elements they focus on. 
To facilitate both channels, we introduce instruments for humans to seamlessly integrate into their demonstrations.

\p{Communicating Relevant Features}
Task actions often depend on contextual variables such as the position of a target object, the color of a traffic signal, or the speed of a moving obstacle. 
We can enable humans to communicate these variables to the robot by equipping them with the required sensors and interfaces. 
In this work, we provide humans with physical \textit{markers} to specify poses and waypoints relevant to the desired task.
Specifically, we let humans place ArUco markers~\cite{garrido2014automatic} in the environment before providing demonstrations.
These markers then continuously stream their poses $b \in \mathbb{R}^{d_{b}}$ to the robot as the human performs the task.
The ArUco markers have a binary pattern that can be detected by the robot's camera for pose estimation; these markers are also small (one inch in width), lightweight (${\sim}10$ grams), and adhere to various surfaces in the environment.
Consider our running example in \fig{method}: when teaching the robot to pick a coffee cup, humans may attach a marker to its side to indicate where they want to grasp.
The marker poses $b$ directly inform the task actions (i.e. how the human teleoperates the robot). 
Hence, we consider poses $b$ as task-relevant features that the robot learner should extract from its observations and incorporate in its control policy.

While we only use positional markers in our experiments, $b$ can more generally include any variables measured through sensors placed by humans in the environment. For example, users could deploy pressure sensors to communicate the force required to grasp different objects during training.

\p{Communicating Relevant Visual Elements}
Not all features essential for performing the task can be directly communicated using markers. For instance, along with the grasping pose of the coffee cup, the human might also care about the color of an indicator light on the coffee machine.
Of course, we could develop a sensor to measure this new variable --- but it would be much more convenient for the user if they could just \textit{describe} the features of interest.
We therefore enable humans to direct the robot's attention toward relevant visual elements by using \textit{natural language} instructions $l \in L$. 
For example, human teachers may say ``pick up the coffee cup'' and ``look at the light on the coffee machine'' when teaching the robot to make coffee. 
While these instructions do not specify the features explicitly (such as the measured grasping pose) they help the robot understand which aspects of the environment the human focuses on (e.g., the cup and coffee machine) and which they ignore (e.g., other objects like the sugar box).

To connect the human's utterances with visual observations we leverage a language-conditioned video segmentation model, \textit{DEVA}~\cite{cheng2023tracking}.
In practice, DEVA associates the human's verbal prompts with objects in the robot's view, producing bounding boxes $\{\mathcal{B}\}$ around objects the human mentions.
Note that humans can also indicate relevant objects using ArUco markers. 
We therefore harness the markers similarly to language, and give them a dual purpose: in addition to estimating marker pose, we detect objects closest to the marker and retrieve those objects' bounding boxes.
Similar to the human teacher, the robot should focus on the visual elements within the bounding boxes when extracting the features.
We expect that this attention will reduce causal confusion with irrelevant objects and allow the robot to implicitly infer task-relevant features from the demonstration data.

\p{Data Collection}
Overall, we shift the demonstration process so that human teachers can use physical markers to explicitly convey relevant poses, and natural language (or markers) to indicate relevant objects for implicit features.
\fig{method} shows how we integrate these instruments into the learning pipeline. 
We ask humans to place the markers before providing demonstrations (Step 1), and then issue natural language commands as they demonstrate the task (Step 2).
Our experimental data suggests that both instruments are intuitive for humans to deploy. 
In our studies, users required less than $15$ seconds to attach the markers to relevant objects for teaching a coffee-making task (see Section~\ref{sec:study}). 
The robot stores the $(b, l)$ data collected from these instruments alongside the states $x$, images $y$, and actions $u$. 
Thus, the augmented dataset $\mathcal{D}$ contains information about \textit{what} actions to imitate and \textit{why} in the form of $(x, y, u, b, l)$ tuples. 
Once all demonstrations have been collected, humans remove the markers from the environment (Step 3). 
We inpaint these markers from images in the dataset so that the robot does not need to rely on seeing the markers in order to perform the task.
The robot only retains the marker poses it recorded during the offline demonstrations.

This is a significant change from standard imitation learning approaches that learn solely from examples of \textit{what} the robot should do, i.e., just $(x, y, u)$ tuples. 
The additional information $(b, l)$ we collect can potentially help the robot resolve causal confusion when learning from visual inputs.
We next present our model architecture and loss functions to train a causal feature function and robot policy from the augmented data $\mathcal{D}$.

\subsection{Network Architecture}\label{sec:model}

Our proposed network architecture is illustrated in \fig{network}. The robot uses an \textit{encoder network} $f_{\psi}(x, y) = \phi$ to map the $n$-dimensional visual observations $y \in \mathbb{R}^{n}$ into $d$-dimensional features $\phi \in \mathbb{R}^{d}$. 
Based on Proposition 1 in Section~\ref{sec:accelerate}, designers should set the dimensionality of these features to be much lower than that of the observations (i.e., $d << n$) in order to accelerate robot learning. As we will describe later, $d$ also depends on the number of markers or language utterances that the human teacher provides.

In practice, the robot often has multiple camera views of the environment (such as a static camera and an ego-centric camera).
To synthesize these views our architecture includes multiple encoders --- one for each camera --- and then combines the output of these encoders with the robot's proprioceptive state $x \in \mathbb{R}^{m}$.
This combination captures the robot's current observations. 
To provide more context for the robot's actions (and enable the robot to reason over its recent history) we then collate a sequence of $k+1$ states and corresponding features to form $\mathcal{X} = [(x_{t-k}, \phi_{t-k}), \ldots, (x_{t}, \phi_{t})]$.
This collated data is then input to a \textit{policy transformer} $\pi_{\theta}$:
\begin{equation}
    \pi_{\theta}(\mathcal{X}) = a_{t}
\end{equation}
Here subscript $t$ denotes the data recorded at a specific time step in the human's offline demonstrations. 
The policy transformer takes the states and features as input and predicts an action token $a_{t}$ for the latest time step. 
We map this token to a robot action $u_{t}$ using an \textit{action network} $g_{\sigma}(a) = u$.

Aspects of our architecture follow the structure of previous visual imitation learning approaches \cite{brohan2022rt,haldarbaku}.
But --- as we will show --- the key difference is how we employ supplementary inputs $(b, l)$ to align the learned features with the human's true features.
In what follows we introduce the auxiliary networks and losses needed to achieve this alignment.

\subsection{Supervised Learning with CIVIL} \label{sec:supervised}

We now describe our Causal and Intuitive Visual Imitation Learning (CIVIL) algorithm for training the robot's policy and feature networks on the augmented dataset $\mathcal{D}$. 
Our algorithm consists of two training phases as shown in \fig{network}. In the first phase, we leverage human guidance in the form of markers and language to learn a task-relevant feature representation (and a downstream policy). In the second phase, we train the robot to causally extract these features without any human guidance so that the markers and language are not needed at test time.

We begin by outlining the first phase.
Our training dataset includes two sources of information about the task-relevant features $\phi^{*}$. The marker poses $b$ only constitute a subset of these features; the robot should infer the remaining non-positional features based on the relevant objects highlighted by users with markers and language commands $l$.
We capture this distinction by dividing the robot's features $\phi$ into two components --- one for the positional features \textit{explicitly} communicated by the user, and another for the non-positional features that that are \textit{implicitly} learned by the robot:
\begin{equation}
    \phi = [\phi_{explicit}, \phi_{implicit}] \label{eq:feature}
\end{equation}
We learn these components separately using the marker and language inputs described below.

\p{Explicit features}
The marker poses $b$ directly inform the task actions. Therefore, we want the robot's features $\phi$ to include all the information from the markers.
At the same time, we recognize that the intended feature $\phi^*$ may be different than the beacon's position $b$ --- perhaps the human is trying to convey position-related features such as size, shape, or distance.
To capture this correlation between $b$ and $\phi$ we learn $\phi_{explicit}$ such that it is a minimally sufficient representation of $b$. In other words, $\phi_{explicit}$ should not include any extra information than what is needed to capture the marker data. 
Formally, we can make $\phi_{explicit}$ contain all information about $b$ by minimizing the conditional entropy of $b$ given $\phi_{explicit}$.
\begin{equation}
    H(b \mid \phi_{explicit}) = -\mathbb{E}_{(x, y, b)\sim \mathcal{D}} \log p(b \mid \phi_{explicit}) \label{eq:entropy}
\end{equation}
Minimizing $H(b|\phi_{explicit})$ means that when we see $\phi_{explicit}$ the robot can determine the corresponding $b$ vector. 
However, this does not ensure that the explicit features exclude other irrelevant information. 
To prevent this irrelevant data, we must also minimize the conditional entropy $H(\phi_{explicit}|b)$:
\begin{equation} \label{eq:E2}
    H(\phi_{explicit} \mid b) = -\mathbb{E}_{(x, y, b)\sim \mathcal{D}} \log p(\phi_{explicit} \mid b)
\end{equation}

From our information-theoretic analysis we seek to learn $\phi_{explicit}$ so that it minimizes both \eq{entropy} and \eq{E2}.
We practically achieve this by introducing a \textit{marker network} $h(b \mid \phi_{explicit})$ which maps explicit features to marker readings. 
This network functionally represents the conditional probability $p(b \mid \phi_{explicit})$. 
We train the forward marker network along with the encoder network $f_{\psi}$ by minimizing the following loss function based on \eq{entropy}: 
\begin{equation}
    \mathcal{L}_{explicit} = -\mathbb{E}_{(x, y, b)\sim \mathcal{D}} \log h(b \mid f_{\psi}(x, y)_{explicit})  \label{eq:loss_explicit}
\end{equation}
Here $f_{\psi}(x, y)_{explicit} = \phi_{explicit}$ is the portion of features that we use to encode the relevant poses. 
The loss in \eq{loss_explicit} captures half of our analysis, and ensures that the features encode $b$. To prevent the features from encoding unnecessary information and satisfy \eq{E2}, we make $h(\cdot)$ an \textit{invertible function}. 
This design choice means that when we train $h$ to map the explicit features to the corresponding marker positions, we can also map those positions back to the features without adding or losing any information. 
We ensure that $h$ is invertible by configuring all layers to have the same dimensions --- forming a square matrix --- and not adding any non-linear activation layers in between. 
Consequently, we must set $\phi_{explicit}$ to have the same dimensions $d_{b}$ as the marker data $b$.
In our experiments, we model $h$ as an identity function $I_{d_{b}}$.

Note that the explicit features $\phi_{explicit}$ need not just be the marker positions: they can also contain any non-positional information that is correlated to the marker data. 
For example, the explicit features may capture the size and shape of the cup in the camera images because these aspects vary with the cup's position.
These features can be then used by the robot in a variety of ways, e.g., the robot can estimate the distance of the cup based on its size and use the shape to determine where it should be grasped.

\p{Implicit features}
Other than explicitly specifying the relevant positions through markers, the human also indicates relevant objects using natural language prompts $l\in\mathcal{L}$. 
Here we explain how the robot maps these prompts to the implicit features $\phi_{implicit}$ from \eq{feature}.
Our first step is to locate the objects mentioned by the human within the corresponding image $y$. 
We do this by feeding the image $y$ and language prompt $l$ to a DEVA model to obtain a bounding box $\mathcal{B}$ for each mentioned object. 
We also generate bounding boxes for objects that overlap with any markers detected in the image. 
For each $(y, l)$ pair we thus obtain a set $\{\mathcal{B}\}$ that includes bounding boxes of task-relevant objects.

We recognize that the human teacher understands the desired task, and we assume that we can rely on that human to identify the key objects or aspects of the task via language and markers. 
This implies that parts of the image $y$ outside the bounding boxes $\{\mathcal{B}\}$ are likely irrelevant and should be ignored by the robot when extracting features.
To enforce this, we generate masked images $y' \in \mathbb{R}^{n}$ by setting all pixels in $y$ that are not within the bounding boxes as zero, and incorporate these filtered images into the training dataset $\mathcal{D}$.

The masked images $y'$ retain relevant information (e.g., the cup and coffee maker) and discard most of the extraneous details (e.g., clutter on the kitchen table). 
But still, the robot does not explicitly know which features to extract from these images and must \textit{implicitly} learn them based on the actions demonstrated by the human.
In our approach we learn the implicit features by training the encoder network and policy transformer end-to-end to imitate human actions. 
Specifically, we minimize the Kullback–Leibler (KL) divergence between the robot's policy $g_{\sigma} \circ \pi_{\theta}$ and the human's optimal policy $\pi_{\theta^*}$ across the training dataset:
\begin{equation} 
    D_{KL}(\pi_{\theta^*} \mid\mid g_{\sigma}) = -\mathbb{E}_{(x, y', u) \sim \mathcal{D}} \big[\log g_{\sigma}(u \mid a)\big] + C \label{eq:kl}
\end{equation}
Here $a$ is the action token output by the policy transformer $\pi_{\theta}$ given a sequence of states and features $\mathcal{X} = [x_{t-i}, \phi_{t-i}]_{i=0}^{k}$, where the features $\phi_{t-i} = f_{\psi}(x_{t-i}, y'_{t-i})$ are extracted from masked images $y'$.
The constant $C$ represents the entropy of the expert's policy $\pi_{\theta^*}$ which does not depend on the robot's parameters. 
Hence, we can ignore the constant term and obtain the following loss function for the robot's policy:
\begin{equation} 
    \mathcal{L}_{policy}= -\mathbb{E}_{(x, y', u) \sim \mathcal{D}} \left[\log g_{\sigma}(u | \pi_{\theta}(x, f_{\psi}(x, y')))\right] \label{eq:loss_policy}
\end{equation}
Minimizing $\mathcal{L}_{policy}$ trains the policy transformer and action network to imitate the actions in the training dataset, and encourages the encoder network to extract features that facilitate this imitation.
What makes this component of our approach different from prior work is that we extract these features from masked images. 
Remember that the robot does not know the relevant aspects \textit{a priori}, so if we try to infer the underlying features from the raw images $y$ there is a greater change of spurious correlations across the high-dimensional dataset. 
However, when we mask the irrelevant details based on objects referenced by the human, it reduces the entropy of the visual data and the likelihood of learning false associations, enabling the robot to better derive features that explain \textit{why} the human teacher chose their actions. 

To summarize, we mask robot observations based on objects mentioned in the language prompts $l$ to implicitly learn the relevant features with \eq{loss_policy}. In addition, we also use the masked images $y'$ instead of the full images $y$ to explicitly encode the relevant poses $b$ with \eq{loss_explicit}.
This supervised feature extraction and policy learning constitutes the first training phase of CIVIL (i.e., the left side of \fig{network}).

\p{Causal Encoder}
We now describe the second phase of CIVIL shown on the right side of \fig{network}.
So far we have found a way to obtain the task-relevant features during \textit{training}; next, we must consider how the robot can obtain these same features at \textit{test time}.
During demonstrations the human can augment the robot's observations through markers and language, which CIVIL leverages to extract task-relevant and supervised features. But when performing the task autonomously the robot will no longer have this guidance --- so the robot needs to understand how to extract these features from unmasked images $y$ of the environment.

To facilitate this, we freeze the parameters of the \textit{encoder network} $f_{\psi}$ that we trained in the first phase and introduce a new \textit{causal network} $c_{\lambda}$ that will learn to extract the task-relevant features from the unmasked images $y$. 
We train this causal network to map the unmasked images $y$ to the same features as those obtained by the trained encoder network from the corresponding masked images $y'$:
\begin{align}
    \mathcal{L}_{causal} = \sum_{(x, y, y') \in \mathcal{D}} & || f_{\psi}(x, y') - c_{\lambda}(x, y) ||^{2} 
    \label{eq:loss_causal}
\end{align}
By minimizing the loss $\mathcal{L}_{causal}$ we teach the causal network to encode the same task-relevant features that the encoder network learned to extract in the first phase. We expect that this will encourage the causal network to focus on the same regions of the raw images $y$ as those highlighted by the human with their language and markers.

At run time the robot can leverage causal  network $c_{\lambda}$ to filter its camera images.
The robot then passes these filtered images to the policy transformer, which ultimately outputs actions taken by the robot arm. Intuitively, this second training phase removes the dependence on human language or physical markers during online execution.

\p{CIVIL Algorithm}
The steps for training our architecture are listed in Algorithm~\ref{alg:cvil} (and visualized in~\fig{network}). 
The human first places markers in the environment to stream relevant poses and then demonstrates the task while providing natural language prompts to indicate the relevant objects. 
The robot uses the markers and language instructions to obtain bounding boxes for all key objects and mask the irrelevant portions of the robot images. These masked images are added to the training dataset along with the marker readings.
We train our network architecture end-to-end by minimizing the loss $\mathcal{L}_{civil}$ in the first training phase:
\begin{equation}
    \mathcal{L}_{civil} = \mathcal{L}_{policy} + \mathcal{L}_{explicit}
\end{equation}
In the second training phase, we freeze the encoder network and then train the causal network with \eq{loss_causal}.
Overall, the trained causal network models how humans reason over the environment observations, while the trained policy transformer replicates how humans decide the task actions.

\subsection{Implementation} \label{sec:M4}

A public CIVIL repository q available here: \url{https://github.com/CIVIL2025/Implementation}.

During our experiments the robot takes images from both a static third-person view $y_{static}$ and an ego-centric view $y_{ego}$.
Accordingly, we train different encoder networks $f_{\psi_{1}}(x, y_{static}) = \phi_{static}$ and $f_{\psi_{2}}(x, y_{ego}) = \phi_{ego}$ for images from each camera view, and implement two corresponding causal networks $c_{\lambda_{1}}$ and $c_{\lambda_{2}}$.
While $\phi_{static}$ has both explicitly and implicitly learned components as in \eq{feature}, we only extract implicit features from the ego-centric view because it does not always observe the marker positions (i.e., objects move in and out of the ego frame). 
We pass both features $(x, \phi_{static}, \phi_{ego})$ as input to the robot policy.
Before feeding these inputs to the policy transformer $\pi_{\theta}$, we project the states and features into separate tokens of size $128$ and add a sinusoidal positional encoding to each token to indicate its location in the sequence~\cite{vaswani2017attention}. 
The encoders are convolutional neural networks; specifically, we choose ResNet-18 initialized with pre-trained weights~\cite{he2016deep}.
The policy architecture includes a 2-layer transformer encoder followed by a multi-layer perceptron (MLP) action network with two hidden layers.

\begin{algorithm}[t]
    \caption{CIVIL}
    \label{alg:cvil}
    \begin{algorithmic}[1] 
        \State Human adds markers to the environment
        \State Human demonstrates task while giving language prompts: $\mathcal{D} = \{(x, y, u, b, l)\}$
        \State Augment dataset with masked images $\mathcal{D} \leftarrow \mathcal{D} \cup [y']$
        \vspace{0.5em}
        \State Initialize model networks $f_{\psi}$, $\pi_{\theta}$, $g_{\sigma}$, $c_{\lambda}$
        \For {$i \in 1, 2, \ldots$}
            \State Compute $\mathcal{L}_{civil}$ on $\mathcal{D}$
            \State Update $(\psi, \sigma, \theta) \gets (\psi, \sigma, \theta) - \alpha \nabla_{\psi, \sigma, \theta} \mathcal{L}_{civil}$            
        \EndFor
        \vspace{0.5em}
        \State Freeze $f_{\psi}$ network
        \State Augment dataset with play data $\mathcal{D}_{causal} \leftarrow \mathcal{D} \cup D_{play}$
        \vspace{0.5em}
        \For {$j \in 1, 2, \ldots$}
            \State Compute $\mathcal{L}_{causal}$ on $\mathcal{D}_{causal}$
            \State Update $\lambda \gets \lambda - \alpha \nabla_{\lambda} \mathcal{L}_{causal}$            
        \EndFor
        \vspace{0.5em}
        \State \Return Trained networks $c_{\lambda}$, $\pi_{\theta}$, $g_{\sigma}$
    \end{algorithmic}
\end{algorithm}

\p{Play data}
CIVIL enables robots to align their representations with those of the human teacher.
But to generalize these representations to new scenarios the robot may still require variability in the training dataset $\mathcal{D}$. 
For instance, if we train the robot to extract relevant poses for just one location of the coffee cup, it may not be able to accurately determine the poses of the coffee cup in new test configurations.

Fortunately, having instruments for explaining human reasoning enables the robot to cheaply train the causal network without needing more human demonstrations.
When feasible, the robot collects additional language prompts $l$ and relevant features $b$ for new task instances that are outside the initial demonstrations, and stores it with the observations $(y, b, l) \in \mathcal{D}_{play}$.
Note that this is an optional step and does not require humans to demonstrate \textit{what} actions to take. The play data $\mathcal{D}_{play}$ only includes information of the relevant objects and poses (i.e., the \textit{why}).
We combine this play data with the training data $\mathcal{D}$ to create an augmented dataset $\mathcal{D}_{causal} = \mathcal{D} \cup \mathcal{D}_{play}$, and use it to train the causal network by minimizing both $\mathcal{L}_{explicit}$ and $\mathcal{L}_{causal}$. We do not use $\mathcal{D}_{play}$ to train the feature networks or the policy transformer.

In summary, our proposed algorithm leverages markers and verbal prompts to bootstrap the learning process and mitigate causal confusion.
Both types of human inputs contribute to improving the robot's understanding of the human's underlying features, resulting in a compact representation of the robot's visual observations.
In the next two sections, we demonstrate the significance of each input and compare CIVIL to state-of-the-art baselines for offline visual imitation learning.
\section{Simulations} \label{sec:sims}

We start by evaluating CIVIL on simulated tasks. 
Our goal is to test whether the proposed algorithm improves learning efficiency and reduces causal confusion by helping robots align their feature representations with those of a human expert.
Across multiple simulated tasks, we compare performance between CIVIL and state-of-the-art baselines for contexts within and outside of the training distribution. 
Unlike CIVIL, these baselines learn feature embeddings through self-supervised transformations of the robot's images, segmenting known objects, or using pre-trained vision-language features. Below we describe these baselines in more detail:
      
\begin{itemize}

\item \textit{Behavior cloning (BC)}~\cite{jain2019learning}: 
A standard imitation learning approach. BC learns to encode camera images and map them to robot actions by only training the policy based on the human's demonstrated actions. This approach forces the robot to implicitly infer the task-relevant features.

\item \textit{Self-Supervised Features (BYOL)}~\cite{grill2020bootstrap}: A self-supervised framework that learns image representations by mapping different views to the same feature encoding. The alternative views are generated using transformations such as random cropping, flipping, and color jittering.
BYOL learns visual features that are not supervised to align with the human's intention. 
In our experiments we pre-trained a BYOL encoder on the images in the training data as well as the play data, froze it, and then used its self-supervised features to train the downstream policy.    

\item \textit{Object-Oriented Features (VIOLA)}~\cite{zhu2023viola}: 
An approach that encodes images by focusing on objects in the scene. VIOLA uses a pre-trained Region Proposal Network (RPN)~\cite{ren2015faster} to obtain bounding boxes for $k$ observed objects and then extracts object-specific features.
In our simulations we provided VIOLA with perfect detection by giving it the ground-truth bounding boxes of all objects in the environment. We then randomly selected $k=5$ of these objects to extract object features as in the original implementation.
We ensure that these objects include the task-relevant item.
By segmenting known objects, VIOLA learns to ignore background variations. However, the robot still needs to figure out which of the $k$ objects are relevant by training the features and downstream policy to imitate human actions in an end-to-end manner.
Note that --- unlike our approach --- VIOLA requires access to the object bounding boxes even during testing.

\item \textit{Task-Specific Object Features (Task-VIOLA)}~\cite{zhu2023learning}: 
This approach is a variation of VIOLA that enables human teachers to indicate the desired objects by scribbling on the robot's images.
The robot then obtains point clouds of the annotated objects from its depth camera, and extracts features by training the downstream policy to imitate human actions.
We replace object point clouds with image segmentations for a fair comparison with other methods that only use RGB images. Note that (similar to VIOLA) this approach also requires a pre-trained vision-language model during test time to  segment the objects.  

\item \textit{Vision-Language Features (CLIP)}~\cite{radford2021learning}: 
The methods discussed so far only derive features from visual inputs. 
We now include a baseline that learns from both images and language prompts. 
Specifically, we use a CLIP encoder that associates visual concepts with their text descriptions by mapping both inputs to the same feature space. 
CLIP features trained on large text-image datasets are general-purpose and may not directly apply to downstream robot tasks~\cite{radosavovic2023real}.
Therefore, we take a pre-trained ResNet-based CLIP encoder RN50$\times$4 and fine-tune it for our simulation tasks by adding top and bottom adapter layers and then training them with the robot policy as in~\cite{sharma2023lossless}.

\end{itemize} 

Unlike these baselines our \textit{CIVIL} approach leverages human inputs from the demonstration process to supervise a causal feature embedding. In contrast to Task-VIOLA --- which lets humans mark relevant objects on a computer screen --- CIVIL does not need a pre-trained model to segment the objects at run time.

\begin{figure}[t]
	\begin{center}
		\resizebox{0.5\textwidth}{!}{\includegraphics{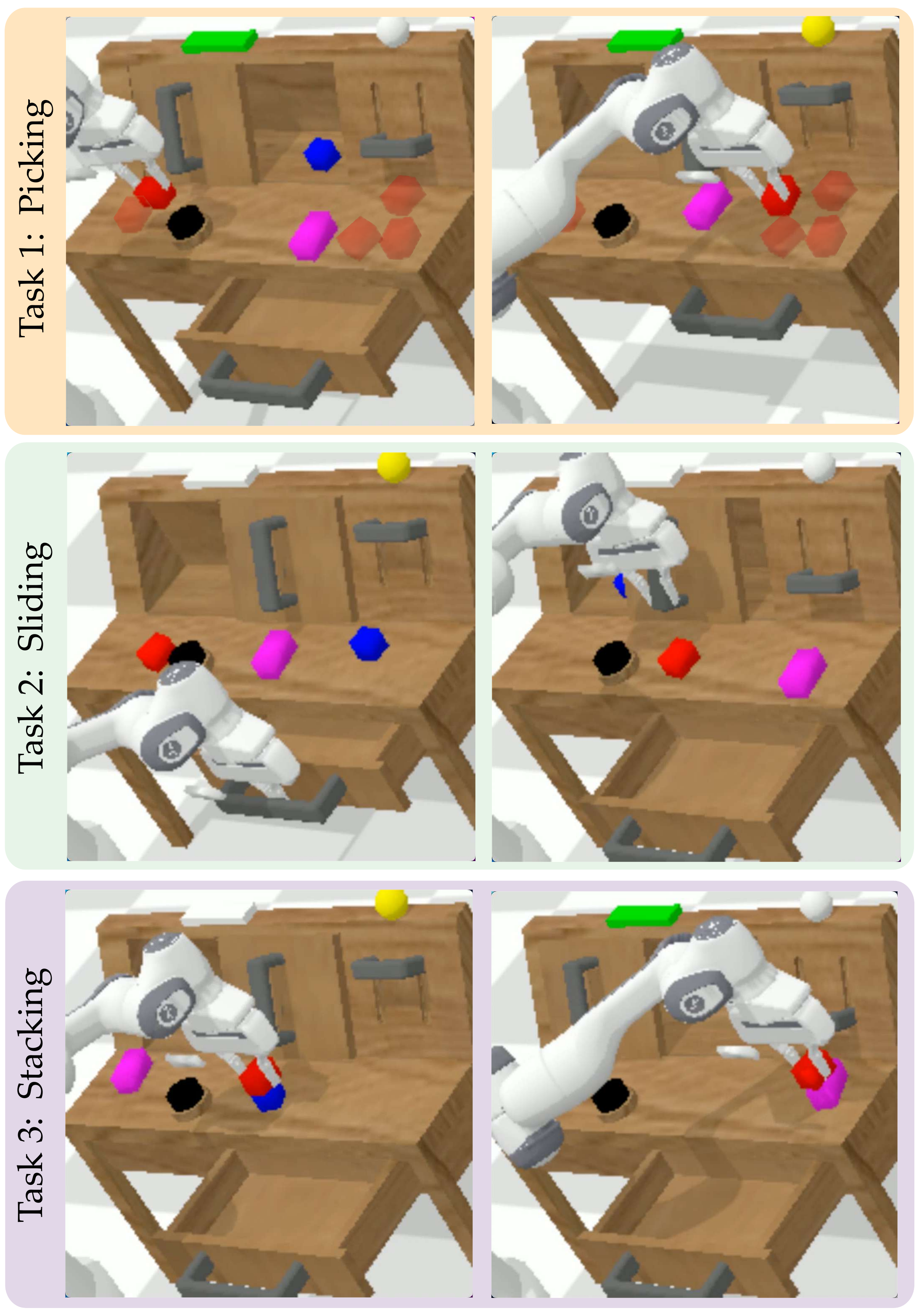}}
		\caption{Manipulation tasks in the CALVIN environment: (1) Picking up a red block. The block is initialized on the left or right side of the table during training. Some of the possible block positions are shown using transparent overlays. (2) Opening the drawer or moving the sliding door based on the light bulb state. The bulb is located in the top right corner and appears yellow when on or white when off. (3) Stacking on the blue or pink block based on the light bulb state and block positions. The task starts with the red block in the robot's gripper and the blue and pink blocks in random positions on the table. In all tasks, the irrelevant objects are also initialized randomly.}
		\label{fig:task}
	\end{center} 
        \vspace{-1.5em}
\end{figure}

\p{Simulation Environment} We trained and evaluated all methods on three tasks within the \textit{CALVIN} environment ~\cite{mees2022calvin} shown in \fig{task}. 
This $3$D environment includes a $7$-DOF Franka Emika Panda robot arm, three differently colored cubes on a workbench, a sliding door, a drawer, a light bulb operated with a control switch, and an LED controlled with a button. 
We randomly initialize these elements during data collection and evaluation.
The demonstrations are either simulated using a pre-trained expert policy~\cite{reuss2024multimodal} or manually collected by an expert teacher. We also had the human expert specify the relevant objects and obtained ground-truth poses of these objects from the simulation environment.

\begin{figure}[t]
	\begin{center}
		\resizebox{0.5\textwidth}{!}{\includegraphics{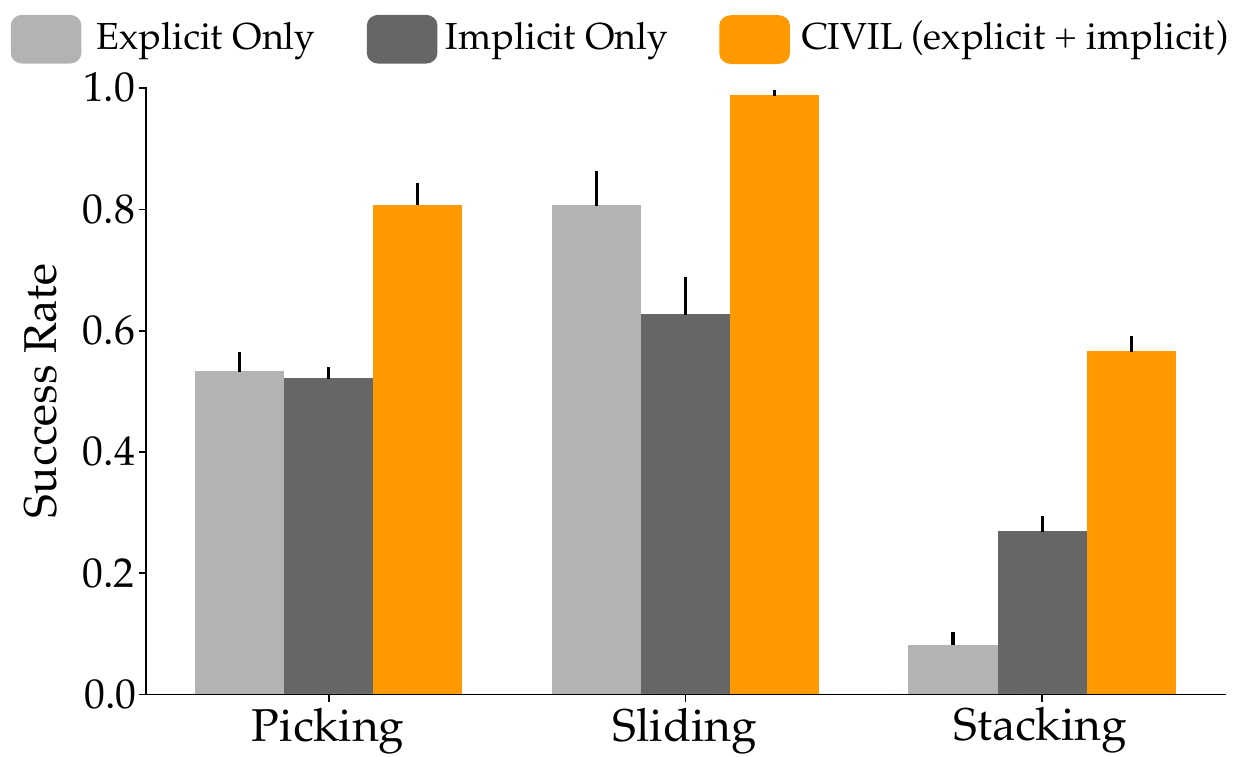}}
		\caption{Results from our ablation study. In \textbf{Explicit} the system is trained on the position data of the marked objects, and in \textbf{Implicit} the system is trained on the masked images. \textbf{CIVIL} takes advantage of the human's explicit and implicit guidance. We find that both components contribute to the overall effectiveness of \textbf{CIVIL}. Each policy is trained with $40$ demonstrations.}
		\label{fig:ablation}
	\end{center} 
        \vspace{-1.5em}
\end{figure}

\begin{figure*}[t]
	\begin{center}
		\includegraphics[width=\linewidth]{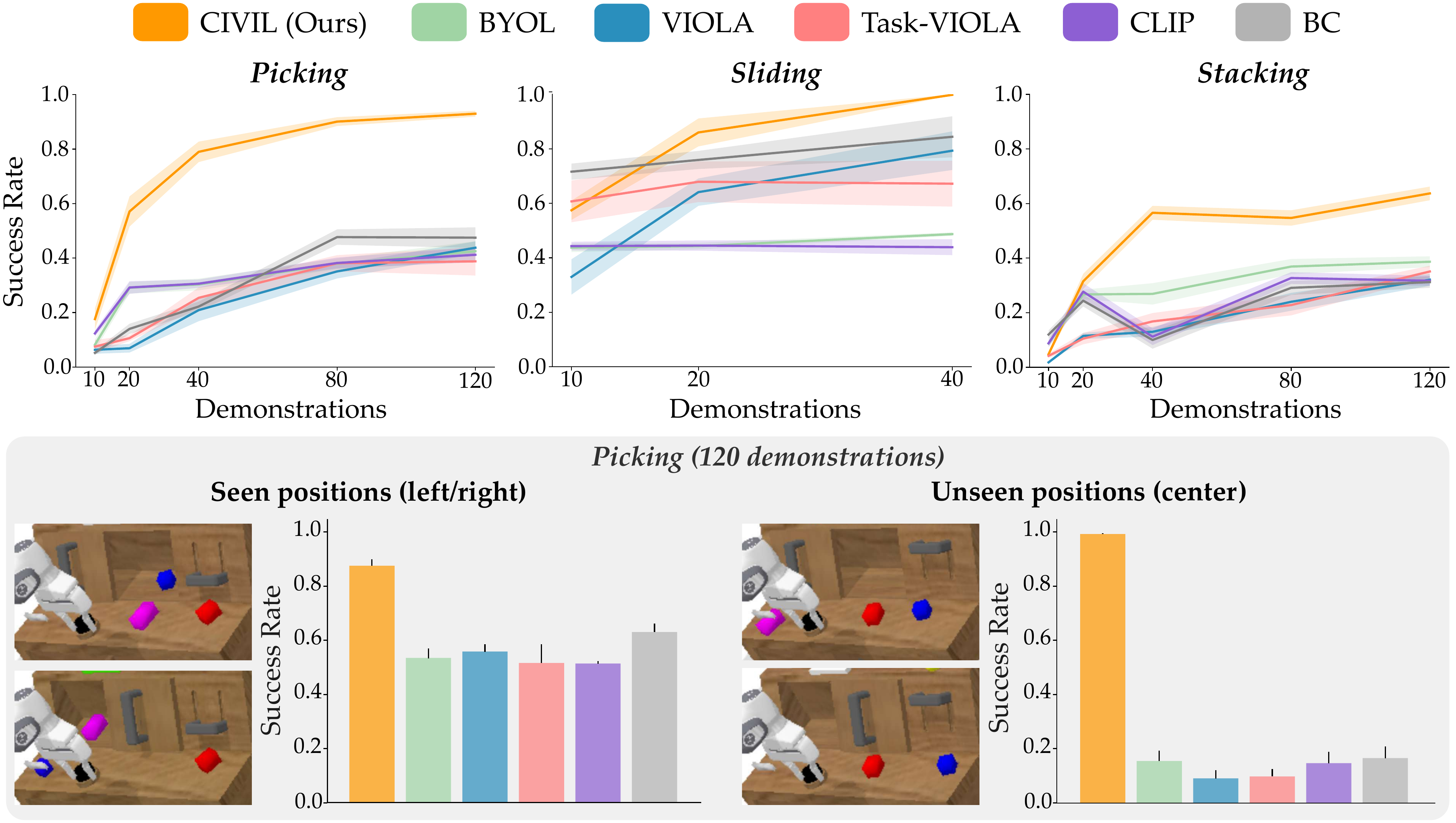}
		\caption{Section~\ref{sec:sims} results for visual imitation learning in the simulated CALVIN environment. We compare our proposed approach (CIVIL) to standard behavior cloning (BC) and baselines that use self-supervised features (BYOL), object-specific features (VIOLA and Task-VIOLA), or vision-language features (CLIP). In the \textit{Picking} task we observe that CIVIL significantly outperforms all baselines in picking up the red block from different positions across the table. Our approach is particularly effective for block positions that are outside the training distribution (i.e., center of the table). This is likely because CIVIL understands what aspects of the image should influence its policy, making the system robust to background clutter or shifting positions. In the \textit{Sliding} task, we find that CIVIL successfully learns to move the drawer or slider based on the state of the light bulb in just $40$ demonstrations. In contrast, baselines that use pre-trained features (BYOL and CLIP) are less precise in detecting the light signal, which reduces their success rate. This suggests that CIVIL can also learn to extract non-positional features (e.g., light bulb state) more efficiently by masking its images based on human-provided language prompts. Lastly, in the \textit{Stacking} task CIVIL leverages both markers and language to extract the positions of the pink and blue blocks as well as the state of the light bulb. This enables the robot to stack the red block more successfully on either the pink or blue block, resulting in a significantly higher success rate. } 
		\label{fig:sim_results}
	\end{center}
        \vspace{-1em}
\end{figure*}

\p{Tasks} We evaluated the methods on the following three tasks in \textit{CALVIN} (see \fig{task}):
\begin{enumerate}
\item \textbf{Picking.}
The robot reaches a red block placed randomly on the table, grasps it, and lifts it to a predefined height. 
This task tests whether the robot can learn features that encode the position of the block and generalize the picking motion to new block positions. 
In the training scenarios, we initialize the red block in a random position on the left or right side of the table (but not in the middle). By contrast, the testing scenarios include block positions across the entire table. 
Here CIVIL measures the pose of the red block during training; accordingly, we expect it to understand that the red block is a key feature, and extrapolate to new block positions at test time.

\textit{Task-relevant objects:} red block

\textit{Marker information:} red block position and orientation

\vspace{0.5ex}

\item \textbf{Sliding.}
The robot arm chooses its behavior based on the state of the light bulb. If the light is on, the robot opens a drawer. If the light is off, the robot instead moves a sliding door. This task tests whether the robot can implicitly extract relevant features that cannot be conveyed directly by positional markers (e.g., whether the light is on or off). 
Our approach receives language prompts that mention the bulb, and leverages these prompts to mask out everything but the relevant objects from its images.
We therefore expect CIVIL to learn the task more efficiently than all baselines except Task-VIOLA, which also receives the segmented image of the light bulb.

\textit{Task-relevant objects:} sliding door, drawer, light bulb

\textit{Marker information:} sliding door and drawer position

\vspace{0.5ex}

\item \textbf{Stacking.} 
In this final task the robot starts with the red block in its gripper and chooses where to place it based on the state of the light bulb. If the light is on, it stacks the red block on a blue block. If the light is off, the red block is stacked on a pink block. The positions of both the blue and pink blocks are initialized randomly. This task tests whether the robot can derive both color-based features (i.e., the light bulb state) that must be \textit{implicitly} learned from masked images as well as positional features (e.g., the block positions) that can be \textit{explicitly} specified with markers. 
Overall, this task combines the challenges of the first two tasks; hence we expect CIVIL to outperform all baselines because the human conveys both relevant poses and objects while training.

\textit{Task-relevant objects:} blue block, pink block, light bulb

\textit{Marker information:} blue and pink block poses

\vspace{0.5ex}

\end{enumerate}

\p{Demonstrations} At each timestep of a task demonstration we record an RGB image $y_{env}$ of size $200 \times 200$ from a static camera that observes the entire manipulation environment, an egocentric RGB image $y_{ego}$ of size $84 \times 84$ from a gripper-mounted camera, an $8$-dimensional robot state $x$, and a $7$-dimensional end-effector action $u$. The state includes $7$ joint angles of the robot arm and a Boolean gripper state. The action is a $6$-dimensional linear and angular velocity and a Boolean gripper actuation.
Additionally, we obtain bounding boxes $\{\mathcal{B}\}$ for all objects in the simulation environment and explicit features $b$ in the form of $6$-dimensional Cartesian poses of relevant objects in that task. 
Each method uses a combination of these inputs to train the feature encoders and robot policy.
For the \textit{Picking} and \textit{Stacking} tasks, we collect an equal amount of play data which includes images, bounding boxes, and relevant poses in randomly initialized scenarios, but it does not include robot states and expert actions.

\p{Training} 
We train all methods for $500$ epochs using the Adam optimizer with a learning rate of $0.0001$ and a scheduler that decreases the rate by a factor of $0.5$ every $100$ training epochs. Our batch size is $128$.
During training, we leave out $10\%$ of the training data and use it as a validation set to evaluate the model after each epoch. 
After training is complete, we save the model instance with the lowest loss on the validation set. 
The validation loss is the mean squared error (MSE) between the expert and model predicted actions.

\p{Ablations}
We compare CIVIL with two ablation variants in the simulation environment to evaluate how the explicit and implicit feature modules contribute to CIVIL's overall performance (\fig{ablation}). 
In the \textit{explicit-only ablation} training is restricted to the first phase, where the policy loss $\mathcal{L}_{\text{policy}}$ is optimized using unmasked image observations, and the explicit supervision loss $\mathcal{L}_{\text{explicit}}$ is applied to marker data. 
By contrast, the \textit{implicit-only ablation} trains the causal network without incorporating $\mathcal{L}_{\text{explicit}}$ during the first training phase. 
For each task, the policies are trained using $40$ demonstrations. 
Observations from the rollouts suggest that the visual markers are particularly useful for recognizing and localizing the target object when it is not yet directly visible in the gripper camera. 
Policies trained with $\mathcal{L}_{\text{explicit}}$ tend to remain closer to the desired trajectory during the early stages of a rollout. 
However, in tasks involving multiple relevant objects such as stacking, the explicit features require more diverse play data to generalize effectively.
The implicit features are designed to capture color or $2$-dimensional segmentation patterns to complement the information not represented in the $3$-dimensional spatial coordinates learned by explicit features. In practice, we found that \textit{implicit-only ablations} utilizing causal networks learned richer representations from gripper images, enabling more precise stacking behaviors. Combining explicit and implicit features helps CIVIL achieves a highest success rate across all three tasks.

\p{Results} Our results are summarized in \fig{sim_results}. Each method is trained on datasets having $10$, $20$, $40$, $80$, and $120$ demonstrations. For statistical robustness, we conduct $10$ independent training and testing runs for each method and dataset size. In each run, we test the trained policy across $100$ randomized task configurations and report the average success rate.

\textit{Picking:} 
For the picking task we found that our CIVIL algorithm outperformed all baselines.
A two-way ANOVA test indicated significant main effects for the choice of method ($F(5, 270)=179.84$, $p<0.001$) and the number of demonstrations ($F(4, 270)=223.08$, $p<0.001$) on the success rate. Post hoc comparisons using Tukey's HSD test found CIVIL to be significantly more effective than the alternatives ($p < 0.01$).

To explore why our approach was more successful than the baselines, we separately examined their performance when the red block was initialized in positions similar to those in the training dataset (i.e., left or right side of the table) and when the red block was placed outside the training distribution (i.e., center of the table).
See \fig{sim_results} (bottom).
When trained with $120$ demonstrations, CIVIL almost always picked up the red block from the center of the table while the baselines had less than a $20\%$ success. 
The difference between the methods was less pronounced when picking the red block from known regions of the table: for these previously seen positions the baselines had a success rate higher than $50\%$, while CIVIL grasped the block in more than $80\%$ of the configurations.
Taken together, these results suggest that the baselines may have overfit to the training distribution, or learned policies that are correlated with the extraneous objects.
By contrast, CIVIL correctly understood \textit{why} the human teacher chose their actions, and learned a policy that reached the red block despite environmental changes and distribution shifts.

\textit{Sliding:} 
In this second task the drawer and slider locations are fixed across all task configurations. 
Instead of focusing on object positions, now the robot needs to learn to condition its behavior on the state of the light bulb (while ignoring distractors within the scene).
Here we observed that CIVIL successfully learned the task after training on just $40$ demonstrations.
The baselines, on the other hand, were unable to achieve the same success rate.
A two-way ANOVA test indicated significant main effects for the choice of method ($F(5, 162)=27.33$, $p<0.001$) and the dataset size ($F(2, 162)=19.79$, $p<0.001$) on task success.

We conducted pairwise comparisons to better understand the differences between methods. Both approaches that used pretrained features performed poorly on this task. A Tukey’s HSD post-hoc test revealed a significant difference in the performance of CIVIL and BYOL ($p < 0.001$) and CIVIL and CLIP ($p < 0.001$). We posit that CLIP underperformed since it is pretrained on real images, which may not adapt well to the simulated environment even after fine-tuning. BYOL is trained on simulation images, but learns features through self-supervision that may fail to emphasize the task-specific light state.
On the other hand, the object-oriented approaches performed better because they focused on a small set of objects including the light bulb. Despite this advantage, both VIOLA ($p < 0.001$) and Task-VIOLA ($p<0.05$) achieved a significantly lower success rate than CIVIL.

Surprisingly, we found that standard behavior cloning performed well in this task. We attribute this result to the small size of its feature space. While BC only extracts one feature token for each camera, the object-centric methods extract two tokens: global and object-specific features. Following our analysis in Section~\ref{sec:accelerate}, a more compact feature space could enable BC to learn more efficiently.
Overall, this simulation result illustrates that by masking images based on language prompts CIVIL is able to extract non-positional features that help it perform the task more successfully.

\textit{Stacking:}
Our final simulation combines the challenges from the first two tasks. Here we observed that CIVIL achieved a significantly higher success rate than all baselines.
A two-way ANOVA revealed significant main effects for method choice ($F(5, 270)=80.31$, $p<0.001$) and demonstration count ($F(4, 270)=163.8$, $p<0.001$). 
Further, post hoc comparisons with Tukey's HSD test indicated that CIVIL was significantly more effective than the baselines ($p < 0.001$).
Since we used the same policy architecture for all the methods, the differences in their success rate were predominantly due to the features they extracted.
This indicates that CIVIL captured both types of task-relevant features more effectively --- the position of the block and the visual state of the light bulb.

\p{Takeaways}
Our simulation results demonstrate that in a cluttered environment with distracting visual elements, CIVIL consistently learns to perform the manipulation tasks from fewer demonstrations as compared to approaches that do not seek to align human and robot representations directly. 
This highlights the benefit of augmenting task demonstrations to convey not just what actions to take but also how to decide on those actions. Specifically, we found that using markers to indicate relevant positions enables robots to generalize to new configurations, and using language prompts to identify and mask-relevant objects enables robots to efficiently learn tasks without being confused by irrelevant items.

What sets CIVIL apart are the additional human inputs we collect as part of the demonstrations and play data. 
Thus far we have shown the benefit of markers and language in a simulated setting and assumed that these instruments are deployed by an expert. 
But how useful are these inputs in real-world tasks, and can these inputs be easily obtained from novice users? 
In the following sections, we conduct real-world experiments and user studies that evaluate whether the performance of our approach holds in practical scenarios where users have a limited time to collect data: placing markers, providing verbal commands, and demonstrating the task.

\begin{figure*}[t]
	\begin{center}
        \includegraphics[width=\linewidth]{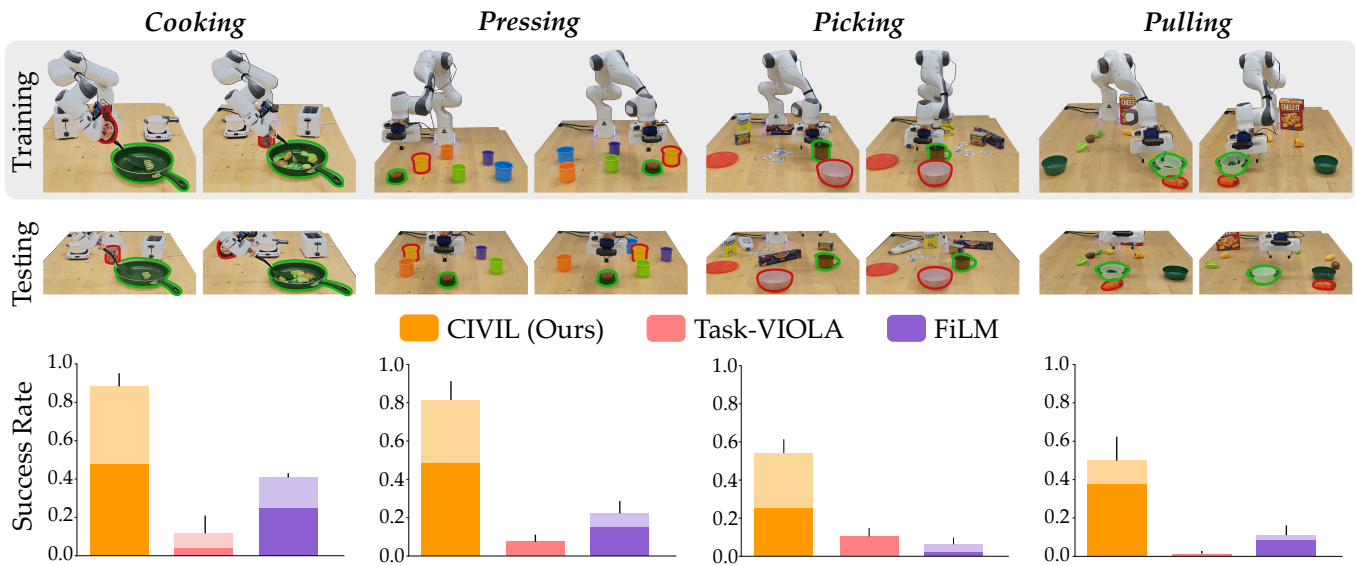}
        \caption{Results for the real-world experiments in Section~\ref{sec:real}.  
        We compare our proposed approach (CIVIL) to object-oriented (Task-VIOLA) and language-conditioned (FiLM) approaches in four manipulation tasks: (1) \textit{cooking} vegetables or meat, (2) \textit{pressing} a red button, (3) \textit{picking} a cup, and (4) \textit{pulling} a bowl to the center of the table. 
        The top row shows examples of training scenarios in each task. We highlight the task-relevant objects in green and one of the distracting objects with a red boundary. During training, the position of the distracting object can be correlated with the relevant object, but this correlation is not present during testing. Example test scenarios are shown in the second row, where the target object appears in an unseen position (except in the cooking task, where the target is fixed).
        The bottom row shows the success rates of the robot arm. We use a darker shade to denote success in seen scenarios and a lighter shade for unseen scenarios. 
        The performance for all approaches drops as the tasks become more complex. In our experiments, cooking was the easiest task as the pan was fixed, while pulling was the most challenging because it involved two target-oriented subtasks, reaching the bowl and bringing it to the center.
        CIVIL achieves a significantly higher success rate than the object-oriented and language-conditioned approaches across all real-world tasks.
        }
        \label{fig:real_task}
	\end{center}
    \vspace{-1em}
\end{figure*}

\section{Real-World Experiments} \label{sec:real}

We now move to a real-world setting where the robot arm performs manipulation tasks on a kitchen table. Compared to the simulation environment, a real scenario presents several challenges: the images are more detailed (e.g., objects have shadows and textures as opposed to a solid color), there is a limited time to collect demonstrations, and the robot may not be able to detect markers and segment images perfectly (i.e., there can be noise in the marker poses and bounding boxes).
Our goal is to test whether CIVIL can still be effective in training robots with noisy inputs and limited data.

In this section we compare our approach to two visual imitation learning baselines: i) \textit{Task-VIOLA}: the method from our simulations that receives segmented images of the task-relevant objects, and ii) \textit{FiLM}~\cite{perez2018film}: a vision-language baseline that uses language prompts to condition its visual features with an affine transformation. We applied this approach instead of CLIP because we found that the features obtained from a pre-trained CLIP encoder did not work well in our real-world tasks during initial testing. Unlike CLIP, FiLM requires language prompts during both training and testing.

\p{Experimental Setup}
We evaluate these methods on a 7-DOF Franka Emika Panda robot arm mounted on a table. The robot uses two Logitech C920 webcams to observe the environment: one serves as a static camera that captures the entire scene, and the other functions as an egocentric camera attached to the end effector of the robot arm. 
We also use a microphone to record verbal instructions during demonstrations. 
To indicate relevant poses, we use $3$D-printed cubes with a width of $20$ millimeters as the physical markers. Five faces of the cube have ArUco tags that uniquely identify that marker, while the sixth face has a reusable adhesive. If the robot detects more than one face of a marker, we take the average of their positions.

\p{Tasks}
We evaluate the methods along four manipulation tasks (also shown in \fig{real_task}):

\begin{enumerate}
    \item \textbf{Cooking.} The robot arm stirs or scoops the contents of a pan with a spatula. If the pan has ``meat,'' the robot scoops it. If the pan has ``vegetables,'' it stirs them. We keep the pan in a fixed location on a table and surround it with objects that can confuse the robot. In particular, during training the robot always sees a tomato can when stirring vegetables or a sauce bottle when scooping meat.
    We test whether the robot can ignore these background objects and learn to act based only on what is in the pan.

    \item \textbf{Pressing.}
    The robot presses a red button on a table that has five cups of different colors. While training, the button is placed on the left or right side of the table, with the yellow cup always located behind the button. During testing, the button can also be in the center and may not be in front of the yellow cup. We test if the robot can avoid being confused by correlations with the yellow cup and push the button regardless of its location.
    
    \item \textbf{Picking.}
    The robot picks up a cup from multiple locations on the table. The robot also sees other objects like a bowl, a spam can, a pasta box, a bleach bottle, and sugar packs. During training, we position the cup on the right side or in the center of the table with the bowl always in front of the cup. However, the cup can be on the left side or at any intermediate location during testing, and the bowl may not be in front of it. 
    As in the previous task, we test whether the robot can avoid being confused by the bowl and generalize to unseen cup positions. 
    
    \item \textbf{Pulling.}
    The robot pulls a bowl to the center of the table. During training, the bowl contains a plastic eggplant and is always placed behind a plastic carrot. There are also other vegetables scattered on the table. However, the eggplant can be in a different container than the bowl during testing. Similar to the previous task, the robot only sees the target object (i.e., bowl) on the left or right side of the table in the training data, but the testing scenarios also include intermediate positions. We test if the robot can learn to focus only on the bowl and not its contents, and generalize to new object positions.
    
\end{enumerate}

\noindent \textbf{Demonstrations} 
The training demonstrations are provided by an expert human using a Logitech joystick. Before performing the task, the expert attaches markers to the target objects. 
During each demonstration, the robot records the static and egocentric RGB images $y_{env}$, $y_{ego}$ which are resized to $200 \times 200$, the $8$-dimensional robot state $x$ which includes $7$ joint angles and one binary gripper state, and a $8$-dimensional action $u$. The action is a $7$-dimensional joint velocity and a binary gripper action.
The robot also tracks the positions $b$ of the markers with the static camera. However, the markers are not detected in every frame so we only obtain marker poses for a subset of the images collected during demonstrations.
The demonstrator provides verbal instructions $l$ (e.g., scoop the meat or stir the vegetables) which are recorded with a microphone and then transcribed to text by a speech recognition model~\cite{radford2023robust}. We use an open-world video segmentation model \textit{DEVA}~\cite{cheng2023tracking} to obtain bounding boxes $\{\mathcal{B}\}$ from text prompts.
Lastly, we in-paint the markers from the images using OpenCV's inpainting tools.

\p{Results}
Our results are summarized in \fig{real_task}. We trained the methods with $20$ expert demonstrations in each task except button pushing, for which we provided $10$ demonstrations. We then tested the methods in several scenarios that reasonably covered all distinct object configurations in each task. Specifically, for tasks numbered $1$ to $4$, we had $40$, $9$, $16$, and $24$ test scenarios, respectively. We measured success based on whether the robot completed the intended task correctly. For instance, if the robot gripper touched anywhere on the button in the pressing task, it was recorded as a success. But if the robot missed the button, it was a failure. The success rates are averaged over $3$ training and evaluation runs. 

We test on both seen and unseen scenarios.
Here we clarify that positions of the irrelevant objects are randomized during testing, and thus no test scenario is exactly the same as a training example. \textit{Seen} therefore refers to contexts where the relevant object (e.g., the button) is in a region of the table where the robot had observed that object at least once during training, while \textit{unseen} refers to the relevant object being in a completely new region. 

The real robot arm performed the tasks more successfully when trained using CIVIL than with the object-oriented or language-conditioned baselines. Our approach performed particularly well in unseen test scenarios, indicating that the robot learned to semantically map the its images into task-relevant and human-aligned features. This result again highlights the benefit of intuitively supervising the robot's features with markers and language, and shows that CIVIL can work well even in real settings where the markers may not be detected at every timestep.
However, contrary to our expectations, FiLM performed considerably better than Task-VIOLA.
Most surprisingly, despite having segmented images of the target object, Task-VIOLA had a less than $15\%$ success rate across all tasks.
We suppose the following two reasons for its poor performance. First, in addition to the object-specific features, Task-VIOLA also extracts global features from the unsegmented image that can contain irrelevant information. As a result, the policy must learn to discard this information implicitly, which is challenging to do given just $20$ demonstrations. Second, Task-VIOLA requires an online object-segmentation approach that may not work perfectly in practice. We found that while it was possible to obtain accurate bounding boxes during the offline training, the robot failed to detect the objects online, especially when they came into contact with the robot's gripper. On the other hand, CIVIL only requires object masks during training and thus does not face the same challenge with online image segmentation.

Overall, our real-world experiments show that given the same number of demonstrations, robots can learn the task more efficiently when supported with markers that specify relevant poses and language prompts that mention relevant objects. However, in practical settings, users may have limited time to provide both demonstrations and the additional inputs. The time required to attach markers and give play data may therefore reduce the number of demonstrations that users can collect. Another factor is that our experiments involved expert teachers who were familiar with placing the markers and giving verbal commands while teleoperating the robot. In the next section, we present a user study that explores whether novice users can do the same under fixed time constraints.
\section{User Study}\label{sec:study}

Now that we have evaluated our algorithm in simulated and real-world tasks with examples provided by a human expert, we will assess whether it is also easy for everyday humans to convey their reasoning while demonstrating the task.
Specifically, we conduct a study to determine if users can intuitively place markers and seamlessly issue language prompts without impacting the quality of their demonstrations. 
We acknowledge that deploying these inputs requires additional time, which may reduce the time left for users to provide examples.
Hence, we also evaluate whether our high-level insight of communicating the key features (\textit{why}) along with the demonstrations (\textit{what}) is practically advantageous when users have a fixed amount of time to teach the robot.
We compare our approach to the behavior cloning (BC) baseline introduced in simulations.
The difference between these methods captures our proposed re-framing of imitation learning: BC learns only from what the human does, while CIVIL enables the human to also convey why they are showing those actions (i.e., the human conveys which robot and environment features their policy depends upon).

\begin{figure*}[t]
	\begin{center}
        \includegraphics[width=\linewidth]{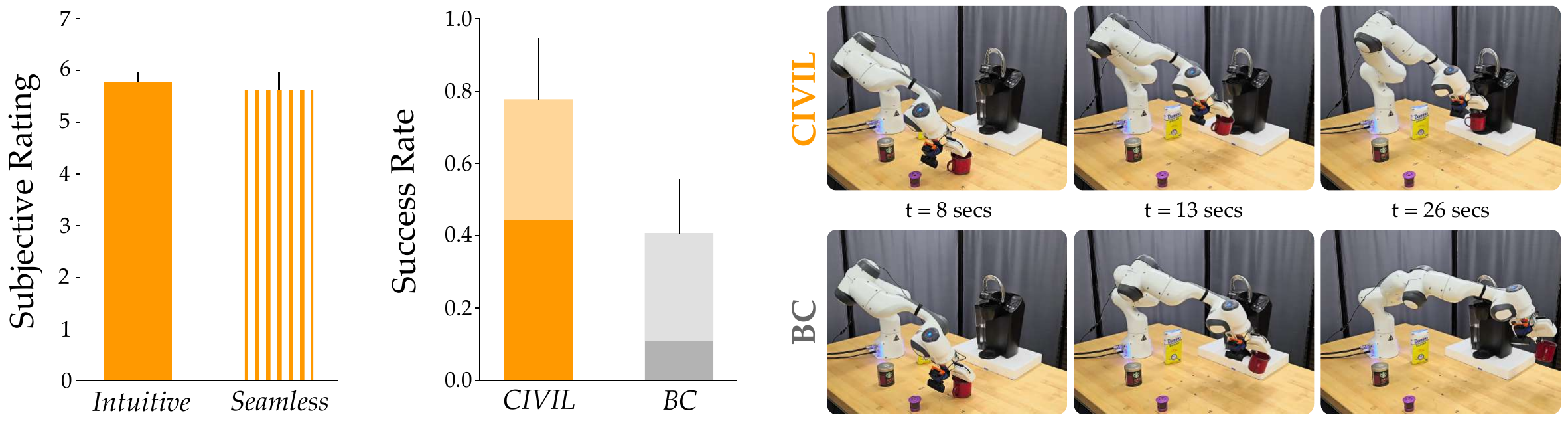}
		\caption{User study results from Section~\ref{sec:study}. (Left) Subjective ratings on a $7$-point Likert scale, where higher values indicate agreement with the statements in Table~\ref{tab:questions}. Users subjectively perceived the process of placing markers and explaining actions to be \textit{intuitive}. Users also reported that they were able to \textit{seamlessly} include these additional steps into their teaching process. (Middle) Objectively, augmenting the demonstrations with feature supervision (CIVIL) led to a significantly better performance than simply providing more action demonstrations to the robot (BC). Here we use a darker shade to denote success in releasing the cup without accidentally toppling it over. (Right) We show an example rollout of both approaches. In this example CIVIL accurately takes the cup to the coffee machine after picking it up, while BC mistakenly takes the cup to the wrong location.}
		\label{fig:study}
	\end{center}
    \vspace{-1.5em}
\end{figure*}

\p{Experimental Setup and Task}
We use the same robot arm and camera setup as in our real-world experiments but choose a new task, \textit{Placing}, for the user study (see \fig{study}). 
In this task the robot has to pick up a cup and place it under a coffee machine. The cup is always initialized in the same position while the machine can be moved along the edge of the table. There are three other objects randomly placed on the table: a coffee pod, a sugar box, and a coffee jar. 
During training the coffee machine only appears in two positions: the nearest and farthest locations along the edge, but during testing it can also be at intermediate locations.

\p{Participants and Procedure}
We recruited $10$ participants ($1$ female, average age $24.4{\pm}3.7$) from Virginia Tech's student population. Participants received monetary compensation for their time and provided informed written consent according to university guidelines (IRB \#23-1237).

At the beginning of the study we showed participants a video of an expert demonstration and gave them $5$ minutes to practice teleoperating the robot using the joystick. During this practice session we also instructed participants on how to attach markers and give language prompts. In particular, we told users that markers should be attached to objects of interest such that they are visible to the robot's camera.
Our study followed a within-subjects design where participants provided data in two rounds, once with markers and language, and once without the additional inputs. \textit{In each round, users had $5$ total minutes to provide as many demonstrations as possible. This included the time required to place markers and collect any play data.}
It is important to note that the markers only need to be attached once at the beginning of one round, and they can remain in place throughout all subsequent demonstrations. Therefore, attaching the markers does not add significantly to the total time required or the human effort involved.
After the trials participants answered a survey (see Table~\ref{tab:questions}) to rate their teaching experiences.
We counterbalanced the order so that half of the participants worked with CIVIL first, and the other half started with BC.

\begin{table}[t!]
\centering
\caption{Survey with two 7-point Likert scales for assessing the intuitiveness of using markers and language, and the ease of incorporating these inputs alongside task demonstrations.}
\label{tab:questions}
\resizebox{\linewidth}{!}{%
\begin{tabular}{ll}
\hline
&\textbf{Intuitive:}\\
\hline
- & Using markers and language feels intuitive and makes sense.\\
- & Using markers and language does not seem intuitive to me.\\ 
- & I understand where to place the markers to help the robot learn.\\
- & I do not understand where I should place the markers.\\
- & I know what verbal instructions will help the robot learn. \\
- & I am unsure what verbal instruction would help the robot learn.\\
\hline
&\textbf{Seamless:}\\
\hline
- & The markers did not interfere while I was performing the task.\\
- & The markers got in my way when I was trying to perform the task.\\
- & Speaking verbal instructions while giving demonstrations did not\\ 
& interfere with my ability to perform the task effectively.\\
- & I was unable to provide effective and accurate demonstrations\\ 
& because I had to give verbal instructions at the same time.\\
\hline
\end{tabular}%
}
\end{table}

\p{Dependent Variables}
To assess the ease of deploying our approach we consider two subjective attributes: \textit{Intuitive} and \textit{Seamless}. 
We measure these attributes through the 7-point Likert scale survey shown in Table~\ref{tab:questions}. 
Users respond to each item in this survey with an agreement rating from $1$ to $7$, where $1$ is strongly disagree and $7$ is strongly agree.
Higher ratings indicate participants found it intuitive to use markers and language, and they could seamlessly integrate these inputs into their demonstrations.
We evaluate the robot's objective performance through the success rate of the learned policy.

\p{Hypothesis}
We made the following hypothesis:
\begin{quote}

\p{H1} \textit{Users will find teaching robots with CIVIL (i.e., showing demonstrations with markers and language) to be just as intuitive and seamless as providing demonstrations for standard BC.}

\p{H2} \textit{Given the same amount of training time, robots using CIVIL will perform the task more successfully than robots with standard BC.}

\end{quote}

\p{Training and Testing}
In a time window of $5$ minutes users provided an average of ${\sim}11$ demonstrations without any additional inputs, and ${\sim}9$ demonstrations when also working with markers and language.
We aggregated the data provided by users into two datasets: i) $\mathcal{D}_{BC}$ which includes the states $x$, images $y$, and actions $u$ from the baseline round, and ii) $\mathcal{D}_{CIVIL}$ which includes marker readings $b$ and language prompts $l$ along with the $(x, y, u)$ samples from our proposed round. 
We also processed the user's language commands to extract the relevant objects. Specifically, we computed the cosine similarity between the text transcribed by Whisper~\cite{radford2023robust} and a pre-defined library containing descriptions for all objects in the environment. For example, ``coffee machine'' and ``black Keurig'' were mapped to ``black coffee maker''.

For testing, we first randomly sampled $15$ demonstrations from $\mathcal{D}_{BC}$ and $13$ from $D_{CIVIL}$ --- amounting to 7 minutes of data --- to train the respective methods. 
We then rolled out the trained models in $9$ scenarios, which included $6$ configurations where the coffee machine was near the closest or farthest point along the table, and $3$ contexts where the coffee machine was in an unseen center position. Other objects were positioned randomly in each scenario. We averaged our final results over $3$ end-to-end runs.

\p{Results}
\fig{study} summarizes our study outcomes.

Overall, users reported that they found it intuitive to deploy markers and speak language commands while teleoperating the robot.
To evaluate their subjective responses, we combined ratings for the survey questions into two scores: one for intuitive and one for seamless.
T-tests indicated that the average user scores for the \textit{Intuitive} ($t(9)=12.99$, $p<0.001$) and \textit{Seamless} ($t(9)=4.34$, $p<0.001$) scales were significantly higher than the neutral score of $4$.
We note that we did not physically show users how to attach markers; we only gave them verbal instructions during the practice round.
Therefore, this result indicates that it was easy for users to understand, remember, and implement our data collection procedure. It also supports our hypotheses \textbf{H1}.
However, we caveat this result with the awareness that $8$ of our $10$ participants stated they had previously interacted with robots, which may have helped them comprehend how robots learn from visual observations and provide more informed training data. 

Given that users understand how to use markers and language, we now explore whether it is worthwhile for them to invest time in providing these inputs when demonstrating the task. 
We observed that robots that were trained with CIVIL learned to grasp the cup and bring it to the coffee machine with a success rate higher than $77\%$ (see the right side of \fig{study}). By contrast, the BC baseline's success rate was about $40\%$, despite receiving more demonstrations than CIVIL. 
This suggests that robots trained without knowledge of the key task features may not realize the human's intent and can be causally confused by the random placement of surrounding clutter in the test scenarios.
It also supports our hypothesis \textbf{H2} and highlights the advantage of a human teacher who conveys the key features (\textit{why to do it}) instead of simply providing more demonstrations of their desired behavior (\textit{what to do}).

Lastly, when taking a closer look at where our approach was superior to standard behavior cloning, we found that CIVIL was significantly more successful in picking up and releasing the cup than BC. For instance, the robot failed to pick up the cup in only $15\%$ of the test scenarios when using CIVIL, compared to $48\%$ when trained with BC. Also, when the robot did manage to pick the cup and take it to the coffee machine, CIVIL was $30\%$ more successful than BC in releasing the cup and moving out without knocking it over. 
Both these instances represent key states in the task where the robot needs to be the most accurate.
This is where training with marker readings helps CIVIL to be more precise than conventional approaches that rely on the robot to extract such positional features without any human guidance.
We also hypothesize that the expressiveness of natural language commands helped non-expert users.
When the robot purely learns from motion demonstrations, any errors in these movements can lead to confusion.
But when the robot reasons over the associated language and markers, CIVIL enables the robot to correctly parse the relevant features, even if the human's motions are imperfect.

In summary, our user study underscores that CIVIL enhances robot learning not by obtaining more data from humans but by providing context to their data. 
We find that CIVIL can significantly improve the robot's ability to learn and generalize to new tasks with only a few context-rich demonstrations.

\section{Conclusion}

In this paper we tackle the problem of causal confusion in visual imitation learning by proposing a fundamental shift in the way humans provide demonstrations, and then leveraging that augmented data to explain the human's actions.
Given just action demonstrations and high-dimensional visual observations, robots can struggle to autonomously extract the correct feature representations.
Without these representations, we analytically and experimentally show that robots may learn to condition their policies on extraneous or spuriously-correlated data, leading to out-of-distribution failures.
To address this challenge, we propose that humans supplement their action demonstrations with additional cues that reveal their decision-making process.
Specifically, we enable humans to deploy physical markers and utter natural language instructions to intuitively convey task-relevant positions and objects that form part of the desired feature representation. 

Our main technical contribution is a visual imitation algorithm, CIVIL, that leverages the verbal prompts to mask unnecessary details from the robot's images and the physical markers to extract a compact feature representation that encodes relevant positional information.
Our simulations and real-world experiments demonstrate that when we use these features to train the robot's policy it learns the task more efficiently, requiring fewer demonstrations than existing approaches. 
CIVIL also enables robots to generalize to new task configurations that are outside the training distribution, indicating that the robot learns features that effectively capture human reasoning.
A distinct advantage of CIVIL is that the robot does not need markers, language instructions, or pretrained vision models at run time when it  autonomously performing the task.

\p{Limitations and Future Work}
This work is a step towards maximizing what robots can learn from human examples. 
However, our current approach has some limitations. 
For instance, we rely on humans to mark or mention the relevant objects. 
This may lead to errors when teaching tasks that contain several relevant components --- humans could forget an essential object or mistakenly mention an irrelevant object.
Future work should account for such potential human errors to prevent the robot from learning incomplete or non-causal representations. A possible way to mitigate this issue would be to actively remind and interact with users throughout the demonstration process.
Another limitation of our work is that we only use verbal commands to identify the task-relevant objects. However, human instructions often contain additional insights such as qualitative descriptions of the demonstrated action (e.g., ``go \textit{straight} to'' or ``place \textit{carefully} under'').
Leveraging these latent signals can help robots make full use of the human's inputs and further accelerate the learning process.
\section{Declarations}

\p{Funding} This research was supported in part by the NSF (Grant Number $2337884$).

\p{Conflict of Interest} The authors declare that they have no conflicts of interest.

\p{Ethical Statement} All physical experiments that relied on interactions with humans were conducted under university guidelines and followed the protocol of Virginia Tech IRB $\#23$-$1237$.

\p{Author Contribution} Y.D. and R.S. led the algorithm development for visual feature extraction. Y.D. contributed to the use of physical markers. R.S. led the setup of the simulation environment. H.N. led the development of the theoretical background. H.N. and D.L. wrote the first manuscript draft. Y.D. and R.J. ran the simulations and R.S. conducted the physical experiments. S.S. and C.N. provided valuable assistance and support throughout the project. D.L. supervised the project, helped develop the method, and edited the manuscript.

\bibliographystyle{spmpsci}
\bibliography{citations}



\end{document}